\documentclass{article}

\usepackage[final,nonatbib]{neurips_data_2024}
\usepackage[toc,page,header]{appendix}
\usepackage{minitoc}

\usepackage[utf8]{inputenc} 
\usepackage[T1]{fontenc}    
\usepackage{hyperref}       
\usepackage{url}            
\usepackage{booktabs}       
\usepackage{amsfonts}       
\usepackage{nicefrac}       
\usepackage{microtype}      
\usepackage{multirow}
\usepackage{array}
\usepackage{boldline}
\usepackage[table, x11names]{xcolor}
\usepackage{epsfig}
\newcommand{\tabincell}[2]{\begin{tabular}{@{}#1@{}}#2\end{tabular}}

\hypersetup{
  colorlinks,
  citecolor=black,
  linkcolor=black,
  urlcolor=black}

\usepackage{float}

\title{PersonalSum: A User-Subjective Guided Personalized Summarization Dataset for Large Language Models}

\author{%
  Lemei Zhang \quad Peng Liu\thanks{Corresponding author} \quad Marcus Tiedemann Oekland Henriksboe \\ \textbf{Even W. Lauvrak} \quad \textbf{Jon Atle Gulla} \quad \textbf{Heri Ramampiaro} \\
  Department of Computer Science, Norwegian University of Science and Technology\\
  \texttt{\{lemei.zhang, peng.liu, jon.atle.gulla, heri\}@ntnu.no} \\
}

\begin{document}

\maketitle

\begin{abstract}
  With the rapid advancement of Natural Language Processing in recent years, numerous studies have shown that generic summaries generated by Large Language Models (LLMs) can sometimes surpass those annotated by experts, such as journalists, according to human evaluations. However, there is limited research on whether these generic summaries meet the individual needs of ordinary people. The biggest obstacle is the lack of human-annotated datasets from the general public. Existing work on personalized summarization often relies on pseudo datasets created from generic summarization datasets or controllable tasks that focus on specific named entities or other aspects, such as the length and specificity of generated summaries, collected from hypothetical tasks without the annotators' initiative. To bridge this gap, we propose a high-quality, personalized, manually annotated abstractive summarization dataset called PersonalSum. This dataset is the first to investigate whether the focus of public readers differs from the generic summaries generated by LLMs. It includes user profiles, personalized summaries accompanied by source sentences from given articles, and machine-generated generic summaries along with their sources. We investigate several personal signals — entities/topics, plot, and structure of articles—that may affect the generation of personalized summaries using LLMs in a few-shot in-context learning scenario. Our preliminary results and analysis indicate that entities/topics are merely one of the key factors that impact the diverse preferences of users, and personalized summarization remains a significant challenge for existing LLMs. Our dataset and code are available at \url{https://github.com/SmartmediaAI/PersonalSum}.
\end{abstract}

\section{Introduction}
Recent studies have demonstrated significant improvements in generating generic summaries using Large Language Models (LLMs) like GPT-3.5, achieving state-of-the-art, even human-level, performance on standard summarization benchmarks. However, personalized summarization, which condenses text to match user preferences while maintaining relevance and non-redundancy, remains largely unexplored. 

Kukoleva \textit{et al.} \cite{Kukoleva} identified three distinct user reading habits in the news domain: \textbf{Attentive reading}, where users read the full article attentively, focusing on details, \textbf{Selective reading}, where users focus only on interesting fragments; and \textbf{Scanning}, where users absorb only the important ideas. User reading focus and attentive reading time can be achieved by monitoring reading duration and scrolling depth, or asking users to annotate their interested parts explicitly. However, there are no existing publicly available resources for research purposes in this area.

Besides, existing textual summarization datasets still suffer from several limitations: to begin with, most data annotators are predominantly journalists or professional writers in related fields, or they are few in number (e.g. only single-digit annotators), resulting in a lack of representativeness, personalization and diversity of the annotated summaries for the general public. Apart from that, the lack of crucial user information, such as reading time and specific content engagement, limits existing research to generic and controllable summary generation. Including user-specific data could enable more personalized and user-centric studies. Furthermore, the existing summarization dataset does not include the user profile of the annotators, which makes our work the only available data for personalized summarization tasks at this stage. 

To this end, we propose PersonalSum, a high-quality human-annotated dataset for personalized news summarization with multiple attributes. In PersonalSum, each article includes personalized summaries annotated by multiple ordinary users based on their interests, multiple pairs of questions and answers related to news articles, and generic summaries generated by machines with manual proofreading. Both the summaries and the question-and-answer pairs contain source information corresponding to the original text. We validated the personalized nature of the dataset and compared it with machine-generated summaries. Today, where LLMs are widely used for text summarization, collecting personalized summaries driven by user subjectivity is indeed challenging. Furthermore, we cannot verify the effectiveness of manual annotations by comparing their similarity to machine-generated summaries. Our analysis revealed that most machine-generated summaries tend to rephrase the introductory sections of news articles, which may also align with user interests. To ensure the annotation quality, we designed an iterative approach combining human evaluation and LLM outputs. Based on user subjectivity, we preliminarily explored the capability of LLMs to generate personalized summaries under different in-context learning settings. Inspired by the findings from Kukoleva \textit{et al.} \cite{Kukoleva} for determining user interest points, we investigated the impact of entities/topics, plot, and article structure on extracting personalized summaries. We have released the above-mentioned datasets, codes and documents at \url{https://github.com/SmartmediaAI/PersonalSum} under a CC BY-NC 4.0 license.

\section{Related work}
\label{related_work}

\begin{table}
\scriptsize
  \caption{Comparison between PersonalSum and existing popular summarization datasets.}
  \label{sample-table}
  \centering
  \setlength\tabcolsep{3.3pt}
  \begin{tabular}{p{2.2cm}p{0.9cm}<{\centering}p{1.7cm}<{\centering}p{1.5cm}<{\centering}p{1.2cm}<{\centering}p{1.2cm}<{\centering}p{0.9cm}<{\centering}p{1.0cm}<{\centering}p{1.2cm}<{\centering}}
    \toprule
     \multirow{3}{*}{Datasets}  & \multirow{3}{*}{Language} & \multirow{3}{*}{Domain} & \multirow{3}{*}{\#Summaries} & \multicolumn{2}{c}{Construction} & \multirow{3}{=}{\centering User Profile} & \multirow{3}{=}{\centering Summary Source} & \multirow{3}{*}{Personalized} \\
    \cmidrule(r){5-6}
    &  &  &  &  Human Annotation & Multi-annotation & & & \\ \midrule
    \hlineB{1.5}
    \rowcolor{lightgray!50} \multicolumn{9}{c}{Generic Summarization Datasets}\\ \hlineB{1}
    CNN/DM \cite{hermann2015teaching} & English  & News & 311,971 & $\surd$ & $\times$ & $\times$ & $\times$ & $\times$ \\
    XSum \cite{narayan2018don} & English & News & 226,711 &  $\surd$ & $\times$ & $\times$ & $\times$ & $\times$   \\
    NewsRoom \cite{grusky2018newsroom} & English & News & 1,212,740 &  $\surd$ & $\times$ & $\times$ & $\times$ & $\times$  \\
    BigPatent \cite{sharma2019bigpatent} & English & Academic & 1,341,362 & $\times$ & $\times$ & $\times$ & $\times$ & $\times$  \\
    arXiv \cite{cohan2018discourse} & English & Academic & 215,913 & $\times$ & $\times$ & $\times$ & $\times$ & $\times$  \\
    PubMed \cite{cohan2018discourse} & English & Academic & 133,215 & $\times$ & $\times$ & $\times$ & $\times$ & $\times$  \\
    LCSTS \cite{hu2015lcsts} & Chinese & News & 2,400,591 & $\surd$ & $\times$ & $\times$ & $\times$ & $\times$  \\
    WikiHow \cite{koupaee2018wikihow} & English & WikiHow & 230,843 & $\times$ & $\times$ & $\times$ & $\times$ & $\times$  \\ \hlineB{1}
    \rowcolor{lightgray!50} \multicolumn{9}{c}{Controllable Summarization Datasets}\\ \hlineB{1}
    DUC \cite{dang2007document} & English & News & 300 & $\surd$ & $\surd$ & $\times$ & $\times$ & $\times$  \\
    QMSum \cite{zhong2021qmsum} & English & Meetings & 1,808 & $\surd$ & $\surd$ & $\times$ & $\times$ & $\times$  \\
    WikiAsp \cite{hayashi2021wikiasp} & English & Wikipedia & 566,881 & $\times$ & $\surd$ & $\times$ & $\times$ & $\times$  \\
    MACSUM \cite{zhang2023macsum} & English & News\&Meetings & 8333 & $\surd$ & $\surd$ & $\times$ & $\surd$ & $\times$  \\ \hlineB{1}
    \rowcolor{lightgray!50} \multicolumn{9}{c}{Personalized Summarization Datasets}\\ \hlineB{1}
    Amazon Reviews \cite{mcauley2015image} & English & E-commerce & 571,540,000 & $\surd$ & $\surd$ & $\surd$ & $\times$ & $\surd$  \\
    PENS \cite{PENS} & English & News Headline & 20,600 & $\surd$ & $\surd$ & $\times$ & $\times$ & $\surd$  \\
    PersonalSum (ours) & Norwegian & News & 1,816 & $\surd$ & $\surd$ & $\surd$ & $\surd$ & $\surd$  \\
    \bottomrule
  \end{tabular}
\end{table}

The personalized textual summarization task is an important yet challenging area in Natural Language Processing. It aims to generate concise and targeted summaries based on a user's personal preferences and focus. However, due to the lack of publicly available datasets, there has been very little research investigating this problem. Currently, publicly available summarization datasets can be roughly divided into three different categories. The first category involves the generation of generic summaries. These summaries can be either pseudo datasets, created by extracting the abstract of the article\cite{xu2023pre,wysoczanska2024tell,dharan2022personalized} or the highlighted sentences within each paragraph\cite{cheng2023towards,xu2021transformer}, or they can be datasets annotated by professional writers or journalists, leveraging their expertise\cite{xu2023sentiment,li2019towards}. 

However, generic text summarization methods frequently fall short in meeting the specific intentions and needs of individual users. This shortcoming has prompted the development of Controllable Text Summarization (CTS) techniques and an expanding corpus of research dedicated to this area \cite{urlana2023controllable}. The task of CTS focuses on creating summaries of source documents that adhere to various controllable attributes or aspects like summary length, coverage of specific topics and so on \cite{hayashi2021wikiasp,zhang2023macsum,he2020ctrlsum,liu2023benchmarking,dou2020gsum}. However, the primary distinction between controllable summarization datasets and PersonalSum lies in the method of controlling the summaries. The former employs explicitly given attributes to guide the generation of summaries, ensuring that these control factors are clearly present in the dataset. In contrast, the personalized news summary dataset proposed in this paper is based on users annotating points of interest within the articles they read. The personalized factors in this dataset are implicit, making them more closely aligned with user interests as reflected in real-world scenarios.

Another type of dataset for textual summarization (e.g. Amazon Reviews \cite{mcauley2015image}) focuses on generating personalized reviews by considering a series of discrete attributes of a given product, which is often an important part of recommender systems \cite{xu2023pre,cheng2023towards,xu2023sentiment,wysoczanska2024tell,dharan2022personalized,li2019towards,xu2021transformer,chan2020unified,li2019persona,liu2019neural}. Different from generic summarization dataset, it usually contains various personalized information e.g., ratings, user and product IDs, and history text, etc. The difference between personalized review summarization and personalized news summarization is that the former often contains user sentiment information. In specific domains, the vocabulary in user reviews tends to be relatively limited. For example, the vocabulary used to describe a movie or a song is not typically used to describe an electronic product, and vice versa. News summaries, on the other hand, often focus on facts and are expressed in a relatively neutral manner, rarely incorporating personal emotions.

So far, the most similar dataset to this paper is called PENS, which is used for personalized headline generation \cite{PENS}. Headline generation is a special case of abstractive text summarization where the goal is to create a concise, often single-sentence "summary" highlighting one key fact of a news article. In contrast, the summaries in PersonalSum may contain two or more user interested points that reflect different perspectives on the news. Furthermore, in PersonalSum, the sources, a feature absent in the other two personalized summarization datasets, can be utilized for various analyses, such as reflecting the user's information extraction habits, serving as a reference for model interpretability, and extracting user interests. A comparison among representative summarization datasets is shown in Table \ref{sample-table}.

\section{PersonalSum}
\label{person_sum}
In this section, we detail our PersonalSum dataset from the perspective of data preparation and collection, while ensuring annotation quality for personalized summarization. 

\subsection{Dataset construction}
\label{data_collect}
The data collection process requires iterative efforts, primarily involving 3 stages, as detailed below.

\textbf{Stage 1: Construction of generic summaries.} In the initial stage, we carefully selected 465 news articles evenly distributed across 10 categories, provided by Schibsted\footnote{\href{https://schibsted.com/}{https://schibsted.com/}}, an international media company. We then developed a data annotation platform and recruited three Norwegian native-speaking students for this task. The students were instructed to revise the GPT-4 generated summaries, emphasizing language fluency, factual consistency, and coherence with the given article while maintaining the focus of the machine-generated summaries. Meanwhile, the students were also asked to highlight sentences from the given news article indicating sources of the summary. After that, each summary was cross-checked by two distinct students with 100\% internal agreement. Besides, each article was paired with multiple question-answer (QA) sets related to the article content, serving as a component of the quality control process during the second phase.

\textbf{Stage 2: Construction of personalized summaries.} In this stage, we recruited annotators from Amazon Mechanical Turk for personalized summarization tasks. First, we instituted a qualification process to assess the suitability of workers for the annotation task. A questionnaire, encompassing fluency in Norwegian, demographic information, news consumption habits, areas of interest, and gender, was administered, allowing us to filter potential workers based on different criteria. Workers without fluent Norwegian knowledge were filtered out. To ensure that each user has certain historical data for later analysis and to ensure the diversity of annotations per article, we then decided to include three articles in each HIT\footnote{In our task, a HIT is one task assigned to an annotator that contains three qualification tests and three different news articles for annotation.}, with each HIT assigned to at least three different annotators. Detailed instructions were provided to guide workers through the annotation process, emphasizing the creation of concise and informative summaries aligned with the annotator's own preferences and interests, while also providing the source of the summary in the given article. Besides, workers were instructed to choose the correct question-answer pair related to the given article from three options to assess comprehension and validate annotations. HITs without a passing rate of 2/3 on the single-choice questions were automatically rejected. Other automatic filtering rules include that the source should come directly from the article, the summary length should not be less than 50 words, and the task duration should not be less than 5 minutes based on the statistics of the time peoples used per reading session in \cite{Kukoleva}, among others, to pre-control the data quality. Considering the market salary level, annotators were compensated \$6 USD per approved HIT, reflecting an hourly rate of \$18 USD, with a projected completion time of 30 minutes per task.

\textbf{Stage 3: Post quality control.} After checking randomly sampled data, we found that a considerable proportion of the collected summaries did not meet the collection standards. They either lack semantic coherence, are irrelevant to the given source, or are annotated in other languages such as English. Inspired by recent work that leverages LLMs as evaluators \cite{liu2023benchmarking}, we utilized OpenAI GPT-3.5-Turbo with few-shot prompting to evaluate the annotated summaries from the perspectives of \textit{Coherence}, \textit{Consistency}, and \textit{Relevance}. In addition to evaluating the coherence of summaries and news articles based on existing work, we especially evaluated the relevance between summaries and sources. Specifically, we score relevance from 0 to 1 according to the relevant scale between the summary and its source. Annotations with a relevance score of less than 0.8 were set aside for human evaluation. For the remaining annotations with a relevance score greater than 0.8, a random sample of 10 percent was taken for human evaluation to estimate the accuracy of the LLM results. This is a very effective way to control the annotation quality. One concern raised during the post-quality control process is that certain sources include only a part of the content of the summary, rather than the entirety. This situation is commonly seen when the relevant score is greater than 0.8. In this context, we have reached a consensus that as long as the source effectively conveys the main points within the summary, it is considered qualified. After the second phase of the first round, only 668 annotations out of 1,395 met the collection criteria. Based on a spot check of the evaluation results, we found that GPT-3.5-Turbo can achieve 98\% accuracy for data with a relevance score greater than 0.8, and 95\% accuracy for evaluations less than 0.8. We then iterate through stages 2 and 3 until the predefined evaluation is satisfied or we reach the budget limit. The prompt used for evaluation is shown in the supplementary material. We use one-shot prompting to control the format of the returned results.

\subsection{Statistical analysis}
\label{sec:stat}
\begin{figure}
\centering
\epsfig{figure=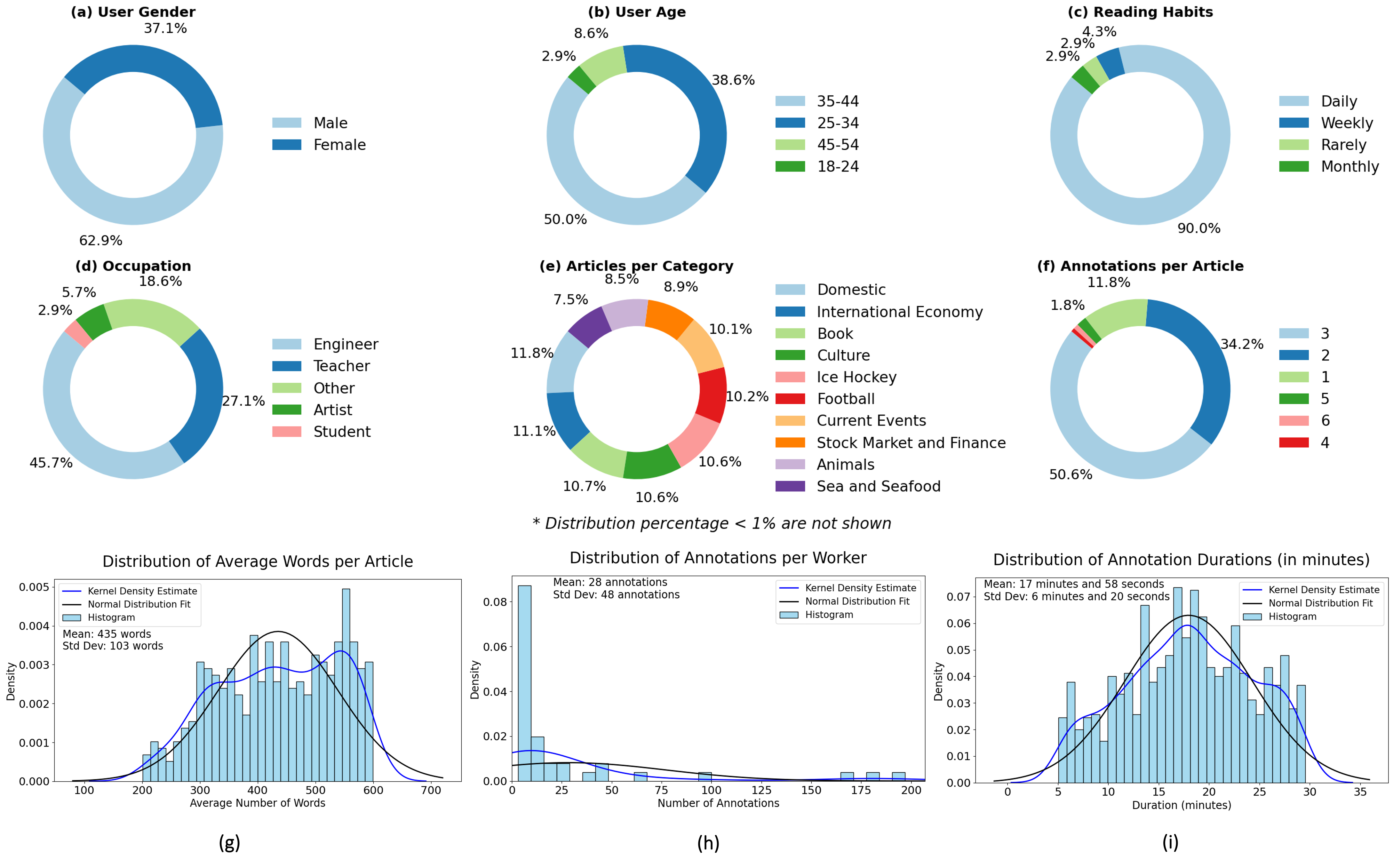, width=386pt,height=225pt}
\caption{(a)-(d) shows annotator demographics, including gender, age, reading habits, and occupation. (e)-(g) cover annotation categories and counts. (h) and (i) display the distribution of qualified summaries per worker and time spent per annotation.} \label{fig:basic_stat}
\end{figure}

\begin{figure}
\centering
\epsfig{figure=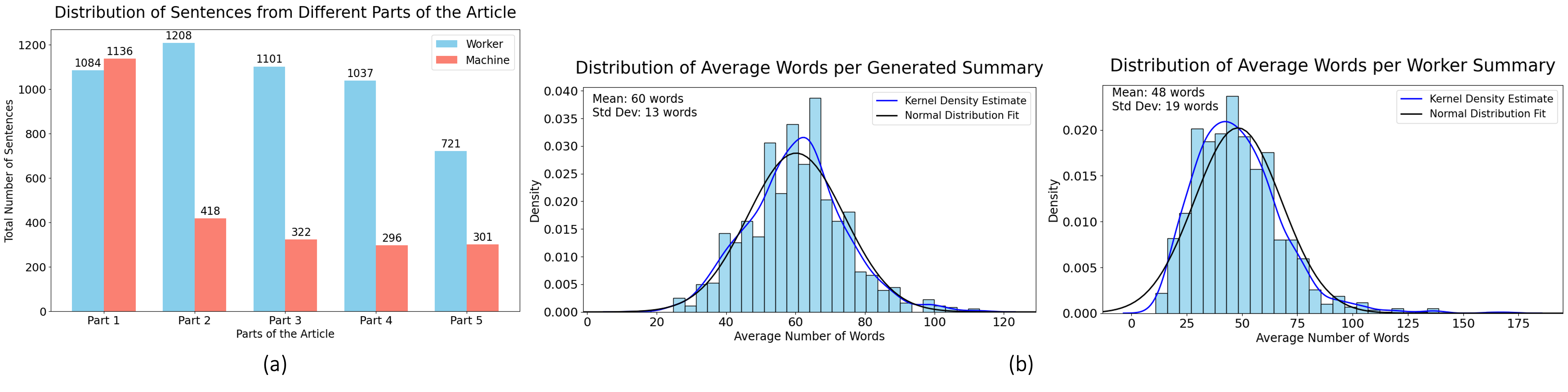, width=387pt,height=102pt}
\caption{(a) The distribution of sources of machine-generated summaries and human-annotated personalized summaries. (b) The distribution of average words per machine-generated summary and human-annotated summary.} \label{fig:difference}
\end{figure}

This section introduces the basic statistics of PersonalSum, covering: 1) annotators, articles, and summaries; 2) the differences between personalized and machine-generated summaries.

After the data collection process described in Section \ref{data_collect}, we collected a total of 1099 personalized summaries from 441 news articles annotated by 39 distinct Amazon Turkers. Figure \ref{fig:basic_stat} (a)-(d) shows the distributions of the annotators based on their gender, age, reading habits and occupation. These distributions indicate that the annotators come from diverse demographic groups, allowing the annotated summaries to represent the perspectives of ordinary people to some extent. Figure \ref{fig:basic_stat} (e)-(g) present basic statistics from the article perspectives. In Figure \ref{fig:basic_stat}(h) and (i),  we present the distributions of qualified annotated summaries per worker and the time consumed per annotation.

Figure \ref{fig:difference} (b) shows that machine-generated summaries are relatively longer than human-annotated ones, but we believe this should not be a major difference since LLMs can generate shorter summaries with different sampling strategies. To illustrate the distribution of summary sources within its given news article, we evenly divide the article into five parts based on the number of sentences, labeling them 1, 2, 3, 4, and 5 from the beginning to the end. Each sentence's source is then identified according to its corresponding part of the article. If multiple sentences originate from the same part, they are counted only once. In Figure \ref{fig:difference} (a), it can be seen that a considerable number of machine-generated summaries originate from the first part of the article. In contrast, the manually annotated summaries are relatively evenly distributed across various parts of the article. This discrepancy may arise because most existing summary datasets use the abstract, typically located at the beginning of the news article, as the ground truth. This observation also indicates that users' focus is diverse and not limited to the initial section of the article. Besides, we analyzed the distribution of different annotation sources for the same article. Out of 441 articles, 52 had only one annotation, 36 had different user summaries originating from the same article part(s), and 353 had annotations from different parts of the article. Notably, the sources for these 36 articles were inconsistent at the sentence level.

\section{Experiments}
\label{sec:exp}
\subsection{Models and evaluation metrics}
Considering the limitations of language and data size in PersonalSum, we tested our dataset on four LLMs across different architectures and model scales, namely OpenAI GPT-3.5 Turbo\footnote{\href{https://platform.openai.com/docs/models/gpt-3-5-turbo}{https://platform.openai.com/docs/models/gpt-3-5-turbo}}, Llama3-instruct\footnote{\href{https://huggingface.co/meta-llama/Meta-Llama-3-70B-Instruct}{https://huggingface.co/meta-llama/Meta-Llama-3-70B-Instruct}}, Google Gemini-1.0-pro\footnote{\href{https://console.cloud.google.com/vertex-ai/publishers/google/model-garden/gemini-pro?pli=1}{https://console.cloud.google.com/vertex-ai/publishers/google/model-garden/gemini-pro?pli=1}}, and 
NorwAI-Mixtral-8x7B-instruct\footnote{\href{https://huggingface.co/NorwAI/NorwAI-Mixtral-8x7B-instruct}{https://huggingface.co/NorwAI/NorwAI-Mixtral-8x7B-instruct}}. All of the selected models are reported to support Norwegian prompting and complex tasks. We benchmarked GPT-3.5-Turbo, Gemini-1.0-pro and NorwAI-Mixtral-8x7B-instruct in zero-shot (for generating generic summaries) and 2/5/10-shot prompting with Norwegian prompts\footnote{We also tested the performance of different models using English prompts, but all results were lower than those using Norwegian prompts.}. For all tested LLMs in this paper, we sampled with a temperature of 0.3, following the work of Wu \textit{et al.}\cite{wu2021recursively}. We evaluated the Llama3-Instruct model using 2-shot prompting and observed that, aside from producing generic summaries, it often generated English summaries for longer Norwegian prompts. This led to significantly lower test results compared to generic summaries. The subpar performance may be due to the limited Norwegian data in the Llama3 pre-training dataset or suboptimal prompt design. Therefore, we chose not to test it with other prompt settings.

We present the performance of PersonalSum on several representative summarization metrics: ROUGE-1/2/L \cite{lin2004rouge}, and BERTScore (F1) \cite{zhang2019bertscore}\footnote{\href{https://huggingface.co/google-bert/bert-base-uncased}{https://huggingface.co/google-bert/bert-base-uncased}}. Inspired by Maynez et al.\cite{maynez2020faithfulness}, who use an entailment score to measure the factuality of generated text compared to human-written text in abstractive summarization, we adopt an entailment model \cite{liu2023nlebench}\footnote{\href{https://huggingface.co/NorGLM/Entailment}{https://huggingface.co/NorGLM/Entailment}} pretrained on NorBERT \cite{kutuzov2021large} to assess the entailment relationship between the generated summaries and the human-annotated personalized summaries in the PersonalSum dataset. Specifically, we consider a generated summary to successfully capture the annotator’s interests if it entails any part of the human-written summary for the corresponding news article. Based on the same work, we calculate the entailment score as the summation of the ratio of support and neutral, as the contradict ratio means that the generated summaries violate the human-written golden summaries. Due to page limits, we only report the best performance of different promptings. The final performances are reported as the average of test results over 5 runs. For the complete evaluation results, please refer to the supplementary materials.

\subsection{Data preprocessing and prompting}
To investigate the impact of three factors, entity/topics, story plots, and article structure, on users' personalized summarization behavior, we utilize GPT-4o\footnote{\href{https://openai.com/index/hello-gpt-4o}{https://openai.com/index/hello-gpt-4o}}, which we found to have superior accuracy in Named Entity Recognition (NER) and the ability to extract simple plot components from given news articles. Specifically, we use GPT-4o to: 1) extract NEs from news articles, summaries, and sources, and 2) extract plot components including event storyline, event cause and event result from news articles. We then compare the user-annotated sources with the article plot to identify which plot components the user highlights in the summary\footnote{We present our prompts of NER and plot extraction in the supplementary materials.}. To investigate article structure, we project the worker summary source distribution from Section \ref{sec:stat} as the worker interested position distribution.

\begin{table}
  \scriptsize
  \caption{2-shot experimental results of different LLMs on PersonalSum. Best results are on bold and the second best results are underlined.}
  \label{2-shot-results-all}
  \centering
  \setlength\tabcolsep{3.9pt}
  \begin{tabular}{p{0.9cm}<{\centering}|p{1.1cm}|p{0.8cm}<{\centering}p{0.8cm}<{\centering}p{0.8cm}<{\centering}p{0.8cm}<{\centering}p{0.8cm}<{\centering}p{1.1cm}<{\centering}p{1.45cm}<{\centering}p{1.25cm}<{\centering}p{0.8cm}<{\centering}}
    \toprule
     Models  & Metrics & Generic & Direct & Entity & Plot & Position & Entity+Plot & Entity+Position & Plot+Position & All \\ \midrule
    \multirow{8}{=}{\centering GPT-3.5 Turbo} & Rouge-1  & \tabincell{c}{37.90\\{\tiny $\pm$14.73}}  & \tabincell{c}{38.01\\{\tiny $\pm$14.82}} & \tabincell{c}{37.56\\{\tiny $\pm$15.23}} & \tabincell{c}{36.90\\{\tiny $\pm$16.25}} & \tabincell{c}{37.93\\{\tiny $\pm$15.38}} & \tabincell{c}{37.93\\{\tiny $\pm$15.36}} & \tabincell{c}{38.03\\{\tiny $\pm$15.10}} & \tabincell{c}{\underline{38.16}\\{\tiny $\pm$15.43}} & \tabincell{c}{\textbf{38.43}\\{\tiny $\pm$15.22}}\\
   \cmidrule(r){2-11}  & Rouge-2 & \tabincell{c}{17.00\\{\tiny $\pm$13.04}}  & \tabincell{c}{17.17\\{\tiny $\pm$13.19}} & \tabincell{c}{16.89\\{\tiny $\pm$13.22}} & \tabincell{c}{16.55\\{\tiny $\pm$13.52}} & \tabincell{c}{17.05\\{\tiny $\pm$13.41}} & \tabincell{c}{17.06\\{\tiny $\pm$13.27}} & \tabincell{c}{\underline{17.27}\\{\tiny $\pm$13.48}} & \tabincell{c}{17.20\\{\tiny $\pm$13.71}} & \tabincell{c}{\textbf{17.47}\\{\tiny $\pm$13.65}}\\
    \cmidrule(r){2-11} & Rouge-L  & \tabincell{c}{26.84\\{\tiny $\pm$13.10}}  & \tabincell{c}{27.16\\{\tiny $\pm$13.13}} & \tabincell{c}{26.85\\{\tiny $\pm$13.31}} & \tabincell{c}{26.28\\{\tiny $\pm$14.05}} & \tabincell{c}{27.15\\{\tiny $\pm$13.65}} & \tabincell{c}{26.96\\{\tiny $\pm$13.49}} & \tabincell{c}{\underline{27.37}\\{\tiny $\pm$13.53}} & \tabincell{c}{27.37\\{\tiny $\pm$13.74}} & \tabincell{c}{\textbf{27.45}\\{\tiny $\pm$13.71}}\\
    \cmidrule(r){2-11} & BERTScore  & \tabincell{c}{75.00\\{\tiny $\pm$5.39}}  & \tabincell{c}{75.16\\{\tiny $\pm$5.30}} & \tabincell{c}{74.76\\{\tiny $\pm$5.72}} & \tabincell{c}{74.64\\{\tiny $\pm$6.14}} & \tabincell{c}{74.98\\{\tiny $\pm$5.79}} & \tabincell{c}{75.00\\{\tiny $\pm$5.64}} & \tabincell{c}{75.02\\{\tiny $\pm$5.62}} & \tabincell{c}{\underline{75.20}\\{\tiny $\pm$5.64}} & \tabincell{c}{\textbf{75.20}\\{\tiny $\pm$5.60}}\\ 
    \midrule[0.3pt]
    \multirow{8}{=}{\centering Gemini 1.0 Pro} & Rouge-1  & \tabincell{c}{35.21\\{\tiny $\pm$13.51}}  & \tabincell{c}{\underline{35.67}\\{\tiny $\pm$14.09}} & \tabincell{c}{35.30\\{\tiny $\pm$13.46}} & \tabincell{c}{35.45\\{\tiny $\pm$13.45}} & \tabincell{c}{35.42\\{\tiny $\pm$13.97}} & \tabincell{c}{35.60\\{\tiny $\pm$13.89}} & \tabincell{c}{\textbf{35.91}\\{\tiny $\pm$14.03}} & \tabincell{c}{35.62\\{\tiny $\pm$14.05}} & \tabincell{c}{35.47\\{\tiny $\pm$13.87}}\\
   \cmidrule(r){2-11}  & Rouge-2 & \tabincell{c}{14.32\\{\tiny $\pm$11.14}}  & \tabincell{c}{\underline{14.76}\\{\tiny $\pm$11.60}} & \tabincell{c}{14.27\\{\tiny $\pm$11.09}} & \tabincell{c}{14.37\\{\tiny $\pm$11.02}} & \tabincell{c}{14.55\\{\tiny $\pm$11.51}} & \tabincell{c}{14.42\\{\tiny $\pm$11.34}} & \tabincell{c}{\textbf{14.88}\\{\tiny $\pm$11.70}} & \tabincell{c}{14.70\\{\tiny $\pm$11.62}} & \tabincell{c}{14.59\\{\tiny $\pm$11.43}}\\
   \cmidrule(r){2-11}  & Rouge-L & \tabincell{c}{25.21\\{\tiny $\pm$11.86}}  & \tabincell{c}{\underline{25.75}\\{\tiny $\pm$12.52}} & \tabincell{c}{25.18\\{\tiny $\pm$11.82}} & \tabincell{c}{25.57\\{\tiny $\pm$11.91}} & \tabincell{c}{25.46\\{\tiny $\pm$12.22}} & \tabincell{c}{25.55\\{\tiny $\pm$12.17}} & \tabincell{c}{\textbf{25.86}\\{\tiny $\pm$12.38}} & \tabincell{c}{25.53\\{\tiny $\pm$12.33}} & \tabincell{c}{25.27\\{\tiny $\pm$11.90}}\\
   \cmidrule(r){2-11} & BERTScore  & \tabincell{c}{74.52\\{\tiny $\pm$5.07}}  & \tabincell{c}{\textbf{74.74}\\{\tiny $\pm$5.24}} & \tabincell{c}{74.42\\{\tiny $\pm$5.02}} & \tabincell{c}{74.56\\{\tiny $\pm$5.01}} & \tabincell{c}{74.49\\{\tiny $\pm$5.28}} & \tabincell{c}{\underline{74.63}\\{\tiny $\pm$5.14}} & \tabincell{c}{74.53\\{\tiny $\pm$5.36}} & \tabincell{c}{74.54\\{\tiny $\pm$5.28}} & \tabincell{c}{74.41\\{\tiny $\pm$5.28}}\\ 
   \midrule[0.3pt]
    \multirow{8}{=}{\centering NorwAI-Mixtral-8x7B-instruct} & Rouge-1  & \tabincell{c}{33.88\\{\tiny $\pm$12.62}}  & \tabincell{c}{34.14\\{\tiny $\pm$13.54}} & \tabincell{c}{34.01\\{\tiny $\pm$13.58}} & \tabincell{c}{33.83\\{\tiny $\pm$13.42}} & \tabincell{c}{33.96\\{\tiny $\pm$13.31}} & \tabincell{c}{34.15\\{\tiny $\pm$13.60}} & \tabincell{c}{33.81\\{\tiny $\pm$13.49}} & \tabincell{c}{\underline{34.24}\\{\tiny $\pm$13.88}} & \tabincell{c}{\textbf{34.29}\\{\tiny $\pm$13.66}}\\
   \cmidrule(r){2-11}  & Rouge-2 & \tabincell{c}{13.36\\{\tiny $\pm$10.43}}  & \tabincell{c}{13.66\\{\tiny $\pm$11.13}} & \tabincell{c}{\underline{13.77}\\{\tiny $\pm$11.00}} & \tabincell{c}{13.56\\{\tiny $\pm$11.13}} & \tabincell{c}{13.69\\{\tiny $\pm$10.95}} & \tabincell{c}{13.75\\{\tiny $\pm$11.05}} & \tabincell{c}{13.62\\{\tiny $\pm$10.96}} & \tabincell{c}{13.77\\{\tiny $\pm$11.26}} & \tabincell{c}{\textbf{13.89}\\{\tiny $\pm$11.03}}\\
   \cmidrule(r){2-11}  & Rouge-L & \tabincell{c}{23.58\\{\tiny $\pm$10.49}}  & \tabincell{c}{24.12\\{\tiny $\pm$11.49}} & \tabincell{c}{24.03\\{\tiny $\pm$11.38}} & \tabincell{c}{23.99\\{\tiny $\pm$11.55}} & \tabincell{c}{24.01\\{\tiny $\pm$11.18}} & \tabincell{c}{\textbf{24.16}\\{\tiny $\pm$11.51}} & \tabincell{c}{24.04\\{\tiny $\pm$11.26}} & \tabincell{c}{24.04\\{\tiny $\pm$11.73}} & \tabincell{c}{\underline{24.13}\\{\tiny $\pm$11.28}}\\
   \cmidrule(r){2-11} & BERTScore  & \tabincell{c}{73.51\\{\tiny $\pm$4.72}}  & \tabincell{c}{73.69\\{\tiny $\pm$4.95}} & \tabincell{c}{73.79\\{\tiny $\pm$4.94}} & \tabincell{c}{73.79\\{\tiny $\pm$4.92}} & \tabincell{c}{\underline{73.84}\\{\tiny $\pm$4.82}} & \tabincell{c}{73.78\\{\tiny $\pm$5.09}} & \tabincell{c}{73.75\\{\tiny $\pm$4.94}} & \tabincell{c}{73.77\\{\tiny $\pm$4.98}} & \tabincell{c}{\textbf{73.95}\\{\tiny $\pm$4.87}}\\
    \bottomrule
  \end{tabular}
\end{table}

\subsection{Evaluation results}
\label{sec:results_all}
We reported experimental results of different models using 2-shot prompting in Table \ref{2-shot-results-all}, where \textit{Generic} refers to generating a summary from the input article without incorporating the user's historical data, with the prompt not accounting for specific factors. \textit{Direct} implies that the prompt does not explicitly include these factors, but it still utilizes the user's previous history. \textit{Plot}, \textit{Entity}, and \textit{Position} indicate that the model prompt is tailored to focus on these particular factors from the user's historical data, with n-shot prompting including this pre-extracted information. The "+" symbol signifies a combination of factors, such as \textit{Entity+Plot}, which considers both. \textit{All} indicates that the prompt instructs the model to account for all relevant factors when generating the summary. We can observe that incorporating the user's historical annotations and personalized factors into the prompt slightly improved the generated results. However, BERTScore struggled to effectively distinguish between different prompt generation results. This may be because most generated summaries convey similar meanings but vary in focus or details. As BERTScore relies on semantic similarity, it loses its advantage in such cases. Besides, through a horizontal comparison of the results generated by 2/5/10-shot prompting, we found that as the number of user's historical annotations in the prompt increases, performance decreases. The possible reason is that the user's annotation data contains scattered features of interest. For example, if a HIT includes both content the user is interested in and content they are not, the user pays attention to different details and reads the article more deeply for the former, as shown in the sources provided. For the latter, the user tends to read only the beginning or end and provides a general summary. 

Through data analysis, we discovered that since articles in each HIT were randomly assigned during data collection, only a small portion of articles within HITs share overlapping entities. As a result, the historical records in the prompt become less relevant to the entity and fail to reflect scenarios where users may focus on specific entities. To better simulate varying user interests in entities, we collected a targeted dataset called Topic-centric PersonalSum, as described in the following section.

\section{Topic-centric PersonalSum}
\label{sec:topic}
\textbf{Experimental design.}  For data collection, we initially grouped articles by identical entities. We observed that many extracted NEs were numbers, publishers, and media brands, which were not indicative of user interest. To enhance the possibility that articles contain shared entities other than invalid ones, we set the minimum number of overlapping NEs in articles within the same HIT to 3. Due to budget constraints, we curated 141 HITs following the steps outlined in Section \ref{data_collect}. Differently, we assigned each HIT to two distinct annotators. Finally, we collected 276 personalized summaries for 72 articles. Among these, 68 articles received at least 2 annotated summaries from distinct annotators, while only 4 articles had a single annotation. 

\begin{table}
  \scriptsize
  \caption{5-shot experimental results of different LLMs on Topic-centric PersonalSum. Best results are on bold and the second best results are underlined.}
  \label{5-shot-results-Topic-centric}
  \centering
  \setlength\tabcolsep{3.9pt}
  \begin{tabular}{p{0.9cm}<{\centering}|p{1.1cm}|p{0.8cm}<{\centering}p{0.8cm}<{\centering}p{0.8cm}<{\centering}p{0.8cm}<{\centering}p{0.8cm}<{\centering}p{1.1cm}<{\centering}p{1.45cm}<{\centering}p{1.25cm}<{\centering}p{0.8cm}<{\centering}}
    \toprule
     Models  & Metrics & Generic & Direct & Entity & Plot & Position & Entity+Plot & Entity+Position & Plot+Position & All \\ \midrule
    \multirow{8}{=}{\centering GPT-3.5 Turbo} & Rouge-1  & \tabincell{c}{37.61\\{\tiny $\pm$13.70}}  & \tabincell{c}{39.75\\{\tiny $\pm$13.86}} & \tabincell{c}{39.68\\{\tiny $\pm$13.79}} & \tabincell{c}{\textbf{40.22}\\{\tiny $\pm$14.55}} & \tabincell{c}{39.39\\{\tiny $\pm$13.74}} & \tabincell{c}{39.10\\{\tiny $\pm$14.19}} & \tabincell{c}{38.68\\{\tiny $\pm$13.60}} & \tabincell{c}{39.31\\{\tiny $\pm$14.37}} & \tabincell{c}{\underline{39.96}\\{\tiny $\pm$13.95}}\\
   \cmidrule(r){2-11}  & Rouge-2 & \tabincell{c}{16.95\\{\tiny $\pm$11.83}}  & \tabincell{c}{18.13\\{\tiny $\pm$12.44}} & \tabincell{c}{18.58\\{\tiny $\pm$12.11}} & \tabincell{c}{\underline{18.62}\\{\tiny $\pm$12.80}} & \tabincell{c}{17.90\\{\tiny $\pm$12.20}} & \tabincell{c}{17.81\\{\tiny $\pm$12.25}} & \tabincell{c}{17.65\\{\tiny $\pm$11.90}} & \tabincell{c}{18.06\\{\tiny $\pm$12.85}} & \tabincell{c}{\textbf{18.71}\\{\tiny $\pm$12.33}}\\
    \cmidrule(r){2-11} & Rouge-L  & \tabincell{c}{26.74\\{\tiny $\pm$12.00}}  & \tabincell{c}{27.86\\{\tiny $\pm$12.34}} & \tabincell{c}{28.08\\{\tiny $\pm$12.37}} & \tabincell{c}{\underline{28.49}\\{\tiny $\pm$13.12}} & \tabincell{c}{27.99\\{\tiny $\pm$12.46}} & \tabincell{c}{27.25\\{\tiny $\pm$12.54}} & \tabincell{c}{27.27\\{\tiny $\pm$11.76}} & \tabincell{c}{27.91\\{\tiny $\pm$12.91}} & \tabincell{c}{\textbf{28.63}\\{\tiny $\pm$12.72}}\\
    \cmidrule(r){2-11} & BERTScore  & \tabincell{c}{75.05\\{\tiny $\pm$5.05}}  & \tabincell{c}{75.64\\{\tiny $\pm$5.04}} & \tabincell{c}{75.57\\{\tiny $\pm$5.23}} & \tabincell{c}{\underline{75.77}\\{\tiny $\pm$5.38}} & \tabincell{c}{75.55\\{\tiny $\pm$5.13}} & \tabincell{c}{75.60\\{\tiny $\pm$5.06}} & \tabincell{c}{75.52\\{\tiny $\pm$5.00}} & \tabincell{c}{75.56\\{\tiny $\pm$5.17}} & \tabincell{c}{\textbf{75.90}\\{\tiny $\pm$5.19}}\\ \midrule[0.3pt]
    \multirow{8}{=}{\centering Gemini 1.0 Pro} & Rouge-1  & \tabincell{c}{35.87\\{\tiny $\pm$13.25}}  & \tabincell{c}{37.35\\{\tiny $\pm$13.20}} & \tabincell{c}{37.54\\{\tiny $\pm$13.49}} & \tabincell{c}{36.96\\{\tiny $\pm$13.39}} & \tabincell{c}{\underline{37.81}\\{\tiny $\pm$13.27}} & \tabincell{c}{\textbf{38.04}\\{\tiny $\pm$13.44}} & \tabincell{c}{36.40\\{\tiny $\pm$13.12}} & \tabincell{c}{37.50\\{\tiny $\pm$13.56}} & \tabincell{c}{37.61\\{\tiny $\pm$13.02}}\\
   \cmidrule(r){2-11}  & Rouge-2 & \tabincell{c}{14.94\\{\tiny $\pm$11.14}}  & \tabincell{c}{15.63\\{\tiny $\pm$10.84}} & \tabincell{c}{\textbf{15.98}\\{\tiny $\pm$11.32}} & \tabincell{c}{15.11\\{\tiny $\pm$10.87}} & \tabincell{c}{\underline{15.95}\\{\tiny $\pm$10.99}} & \tabincell{c}{15.91\\{\tiny $\pm$11.20}} & \tabincell{c}{14.85\\{\tiny $\pm$10.73}} & \tabincell{c}{15.57\\{\tiny $\pm$11.42}} & \tabincell{c}{15.74\\{\tiny $\pm$10.98}}\\
   \cmidrule(r){2-11}  & Rouge-L & \tabincell{c}{25.05\\{\tiny $\pm$11.82}}  & \tabincell{c}{25.96\\{\tiny $\pm$11.68}} & \tabincell{c}{\textbf{26.47}\\{\tiny $\pm$12.30}} & \tabincell{c}{25.48\\{\tiny $\pm$11.86}} & \tabincell{c}{26.09\\{\tiny $\pm$11.70}} & \tabincell{c}{\underline{26.45}\\{\tiny $\pm$12.36}} & \tabincell{c}{25.22\\{\tiny $\pm$11.24}} & \tabincell{c}{26.06\\{\tiny $\pm$12.00}} & \tabincell{c}{25.94\\{\tiny $\pm$11.38}}\\
   \cmidrule(r){2-11} & BERTScore  & \tabincell{c}{74.50\\{\tiny $\pm$5.10}}  & \tabincell{c}{75.02\\{\tiny $\pm$5.13}} & \tabincell{c}{\underline{75.03}\\{\tiny $\pm$5.03}} & \tabincell{c}{74.98\\{\tiny $\pm$5.07}} & \tabincell{c}{74.92\\{\tiny $\pm$5.00}} & \tabincell{c}{\textbf{75.29}\\{\tiny $\pm$5.24}} & \tabincell{c}{74.70\\{\tiny $\pm$4.99}} & \tabincell{c}{75.03\\{\tiny $\pm$5.08}} & \tabincell{c}{74.96\\{\tiny $\pm$5.01}}\\ \midrule[0.3pt]
    \multirow{8}{=}{\centering NorwAI-Mixtral-8x7B-instruct} & Rouge-1  & \tabincell{c}{33.40\\{\tiny $\pm$12.17}}  & \tabincell{c}{\underline{35.29}\\{\tiny $\pm$13.87}} & \tabincell{c}{33.83\\{\tiny $\pm$13.80}} & \tabincell{c}{34.28\\{\tiny $\pm$13.78}} & \tabincell{c}{34.67\\{\tiny $\pm$13.24}} & \tabincell{c}{34.83\\{\tiny $\pm$14.03}} & \tabincell{c}{34.36\\{\tiny $\pm$13.98}} & \tabincell{c}{\textbf{35.75}\\{\tiny $\pm$13.91}} & \tabincell{c}{34.60\\{\tiny $\pm$14.08}}\\
   \cmidrule(r){2-11}  & Rouge-2 & \tabincell{c}{13.05\\{\tiny $\pm$9.57}}  & \tabincell{c}{14.49\\{\tiny $\pm$11.29}} & \tabincell{c}{13.56\\{\tiny $\pm$10.50}} & \tabincell{c}{13.86\\{\tiny $\pm$10.67}} & \tabincell{c}{13.82\\{\tiny $\pm$10.69}} & \tabincell{c}{\underline{14.51}\\{\tiny $\pm$11.10}} & \tabincell{c}{14.18\\{\tiny $\pm$10.76}} & \tabincell{c}{\textbf{14.98}\\{\tiny $\pm$11.01}} & \tabincell{c}{14.23\\{\tiny $\pm$11.00}}\\
   \cmidrule(r){2-11}  & Rouge-L & \tabincell{c}{23.04\\{\tiny $\pm$10.04}}  & \tabincell{c}{24.59\\{\tiny $\pm$11.50}} & \tabincell{c}{23.60\\{\tiny $\pm$10.96}} & \tabincell{c}{24.24\\{\tiny $\pm$11.48}} & \tabincell{c}{23.93\\{\tiny $\pm$11.26}} & \tabincell{c}{\underline{24.82}\\{\tiny $\pm$11.94}} & \tabincell{c}{24.39\\{\tiny $\pm$11.55}} & \tabincell{c}{\textbf{24.99}\\{\tiny $\pm$11.70}} & \tabincell{c}{24.58\\{\tiny $\pm$11.95}}\\
   \cmidrule(r){2-11} & BERTScore  & \tabincell{c}{73.41\\{\tiny $\pm$4.60}}  & \tabincell{c}{74.21\\{\tiny $\pm$4.96}} & \tabincell{c}{73.78\\{\tiny $\pm$5.12}} & \tabincell{c}{73.86\\{\tiny $\pm$5.15}} & \tabincell{c}{73.96\\{\tiny $\pm$4.98}} & \tabincell{c}{\underline{74.27}\\{\tiny $\pm$5.15}} & \tabincell{c}{74.16\\{\tiny $\pm$5.11}} & \tabincell{c}{\textbf{74.38}\\{\tiny $\pm$5.18}} & \tabincell{c}{74.12\\{\tiny $\pm$5.19}}\\
    \bottomrule
  \end{tabular}
\end{table}

\begin{figure}
\centering
\epsfig{figure=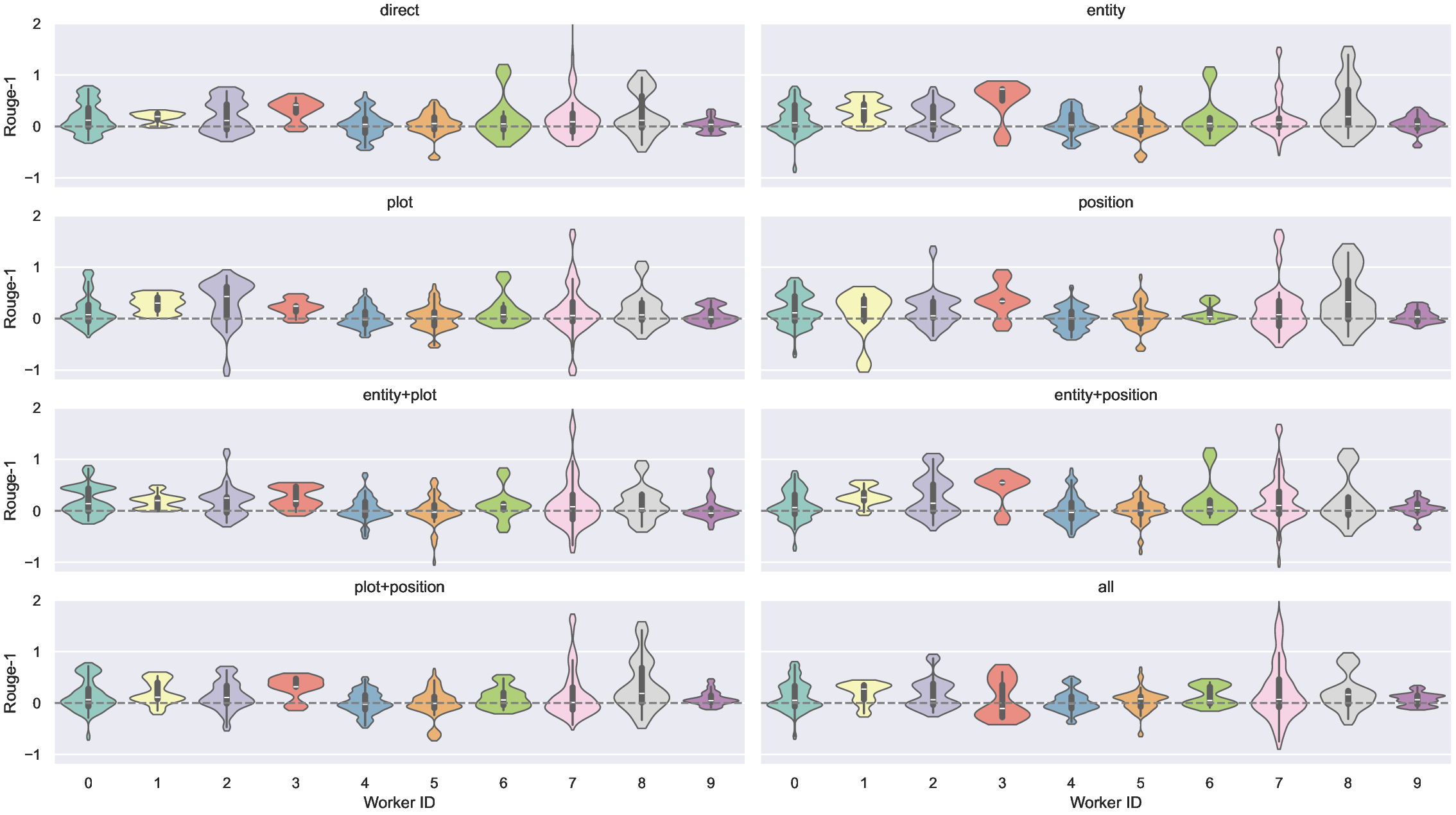, width=400pt, height=240pt}
\caption{Experimental results showing improvements in the ROUGE-1 score from personalized prompting compared to generic summaries using GPT-3.5 Turbo for each worker. The X-axis represents worker IDs, and the Y-axis represents the ROUGE-1 score improvements.} \label{fig:violin}
\end{figure}

\textbf{Results and analysis.} 
We conducted the same experiments as described in Section \ref{sec:exp} on Topic-centric PersonalSum. Experimental results using 5-shot prompting are shown in Table \ref{5-shot-results-Topic-centric}\footnote{Please see supplementary materials for the statistics and the complete experimental results of the Topic-centric PersonalSum.}. We can observe that: First, all personalized results outperform the generic summaries, demonstrating that our data is effectively personalized and captured by different models across various dimensions. Besides, explicitly incorporating diverse factors into the prompt influences the model's output to varying extents. We also observe that 5-shot prompting yields the best results across all models, whereas 10-shot prompting performs worse than 2-shot prompting. This indicates that when generating personalized summaries, it is crucial to balance the number of input user history records. Compared with Section \ref{sec:results_all}, though the best results appear with different few-shot prompting methods, we can still see that an excessive amount of user history data introduces noise to the pre-trained models, adversely affecting the generation outcomes. The superior experimental results on Topic-centric PersonalSum may demonstrate that it exhibits more pronounced user annotation characteristics compared to PersonalSum.

\begin{figure}
\centering
\epsfig{figure=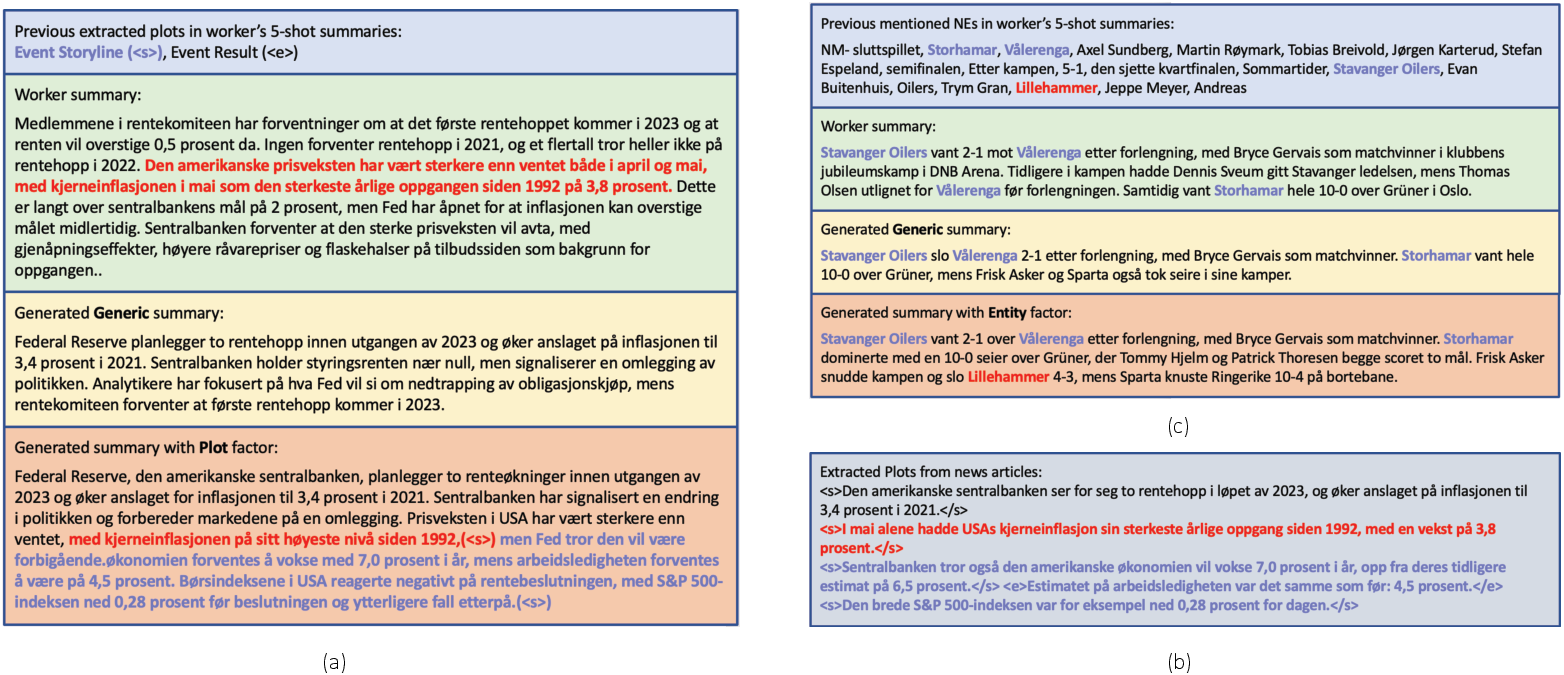, width=400pt, height=178pt}
\vspace{-10pt}
\caption{(a) The plot information concerned in the 5-shot historical annotated summaries of Worker 3, the generic summary, and the summary with the prompt including the annotator's plot information. (b) The article's plot data is extracted by GPT-4o. For clarity, we only include the original information relevant to the generated summaries for Worker 3. (c) The entities that appear in the 5-shot historical annotations of Worker 1, the user-annotated summary, the generic summary, and the summary with the prompt including the annotator's entity details. All generated summaries are from GPT-3.5-Turbo.}  \label{fig:case_study}
\vspace{-8pt}
\end{figure}

To gain a more comprehensive understanding of how different factors impact the models, we investigated the performance from the user side. Figure \ref{fig:violin} shows the experimental results of the improvements in the ROUGE-1 score from personalized prompting compared to generic summaries using GPT-3.5 Turbo for each worker, based on Table \ref{5-shot-results-Topic-centric}. From the differences in improvement, we selected two instances for further analysis: one from worker 1, who showed a higher improvement, and one from worker 3, who showed reduced performance. This analysis aims to have a hint on the impact of various factors. As shown in Figure \ref{fig:case_study} (a) and (b), Worker 3 is interested in event storylines which could be details of the event, and event results. When the model is prompted to generate a summary with a storyline, it introduces additional descriptions about 1992 that align with the user's annotations, compared to the generic summary. However, it also includes descriptions of the S\&P 500 index, which were not annotated by the user. In Figure \ref{fig:case_study} (c), when the model is prompted to generate a summary considering Worker 1's previously annotated NEs, apart from the entities highlighted in purple that match the annotated summary, the model also includes the entity "Lillehammer" which appears in the user's history.

\section{Human Evaluation}
We recruit three well-educated Norwegian native colleague students to evaluate model generated summaries (including generic, direct and all factors) of three models (GPT3.5-turbo, Gemini, and NorwAI-Mixtral-8x7B-instruct) using 5-shot prompting with human written summaries. Considering the time and cost limit, we randomly selected 50 samples from PersonalSum for evaluation. The detailed instructions to the evaluators are shown in Figure \ref{fig:Instructions} in the supplementary materials.

We adopt Fleiss' kappa ($\kappa$) to measure Inter-rater Agreement among the three raters for each evaluation metric and model. The results are shown in Table \ref{tab:human_eva}. The Fleiss' kappa score shows that all evaluation results achieve substantial or almost perfect agreement \cite{landis1977measurement}\footnote{\href{https://www.ncbi.nlm.nih.gov/books/NBK92287/table/executivesummary.t2/?report=objectonly}{https://www.ncbi.nlm.nih.gov/books/NBK92287/table/executivesummary.t2/?report=objectonly}}. From the Table, we can see that while the generic summary preserves much of the content from the user-annotated summary, the summary generated using prompts that explicitly include entities, news plots, or news article structure preference aligns more closely with the user's personalized content needs. In addition, after analyzing the issues present in the generated summaries provided by the raters, we observed that for GPT3.5-Turbo and Gemini1.0-pro, the primary challenges are "2. excessive detail", followed by "1. a focus on different topics" and "4. divergent plot emphasis". In contrast, the primary issues for NorwAI-Mixtral-8x7B-instruct involve "1. a focus on different topics", followed by "4. divergent plot emphasis" and "5. incomplete outputs".

\begin{table}
\scriptsize
\centering
  \caption{Human evaluation results on the quality of personalized summaries generated by LLMs.}
  \label{tab:human_eva}
  \begin{tabular}{m{3.1cm}m{1.3cm}<{\centering}m{1.3cm}<{\centering}m{1.3cm}<{\centering}m{1.3cm}<{\centering}m{1.3cm}<{\centering}m{1.3cm}<{\centering}}
    \toprule
    \multirow{2}{*}{Models} & \multicolumn{3}{c}{\centering Consistency / Fleiss' kappa} & \multicolumn{3}{c}{\centering Coherence / Fleiss' kappa}\\
    \cmidrule(lr){2-4}\cmidrule(lr){5-7}
    & Generic & Direct & All & Generic & Direct & All \\ 
    \midrule
    GPT-3.5 Turbo & 4.03 / 0.96 & 4.02 / 0.91 & \textbf{4.05} / 0.91 & \textbf{4.78} / 0.86 & 4.77 / 0.80 & 4.70 / 0.86 \\
    \hline
    Gemini 1.0 Pro & 3.95 / 0.83 & 4.01 / 0.73 & \textbf{4.03} / 0.86 & 4.69 / 0.82 & \textbf{4.74} / 0.82 & 4.67 / 0.98 \\
    \hline
    NorwAI-Mixtral-8x7B-instruct & 3.81 / 0.82 & 3.87 / 0.71 & \textbf{3.99} / 0.83 & 4.53 / 0.66 & 4.59 / 0.83 & \textbf{4.63} / 0.77 \\
    \bottomrule
  \end{tabular}
\vspace{-5pt}
\end{table}

\section{Discussion and future work}
\label{sec:limit}
From the experimental results on both datasets, we observe the following: 1) Entities play a crucial role in personalized summarization. Despite the similarity or interconnectedness of topics in many HITs articles in the Topic-centric dataset, solely considering the entity factor in few-shot scenarios may not maximize improvement. Plot and article structure could also be among the myriad factors affecting the user's personalized summary. 2) Through the analysis of individual use cases, we found that there are many details in the user's personalized summary, which are often ignored by expert writers and journalists when writing summaries. This is reasonable because professional writers and journalists often extract the main points and salient content to meet the needs of the public. However, we argue that the uniqueness of each individual should not be ignored. (3) Limitations of our work include an insufficient amount of data for model training and the workers were not able to select articles to annotate themselves.


The rich properties of PersonalSum enable its use across various applications, such as evaluating the explainability and factuality of document-grounded question-answering systems and news summarization models, exploring information extraction characteristics, and uncovering the general public's implicit interests. Furthermore, it provides an opportunity to delve into a comparative analysis of machine-generated summaries and those annotated manually. This diverse range of applications underscores the versatility and potential of the dataset in advancing research in these areas.

\section{Ethical Statement}
Prior to the annotation process, all participants were informed about the purpose and intended use of the data they provided. Annotators were given the option to select "prefer not to say" to ensure they feel comfortable sharing personal information. Sensitive data, such as age and occupation, was used solely for statistical purposes and will not be shared or used beyond the scope of the study outlined in this paper. This ensures that participants' identities remain unidentifiable through the annotated data. Also, each users were given all necessary information about the extents of our study, hence ensuring transparency.

To achieve a diverse representation of users, we collected annotations from various age groups, including those under 25 and over 35, since the majority of annotators were aged 25-34. While biases may arise if users' preferences do not align with the assigned news articles, capturing user subjectivity is one of our goals, as it reflects real user behavior and personal perspectives.

We emphasize that this work focuses on exploring the capability of LLMs to generate personalized summaries rather than defining users by their traits or personalities, thus respecting privacy and avoiding unwarranted generalizations. The code and dataset used for data collection and experiments in this paper have been made publicly available on GitHub for reproducibility purposes.

\section*{Acknowledgments}
This work was carried out at the Norwegian Research Center for AI Innovation (NorwAI), funded by the Research Council of Norway through the Centre for Research-based Innovation (SFI) funding scheme, with additional financial support from NorwAI’s partners. 

We extend our gratitude to the reviewers for their valuable feedback. Special thanks to the IDUN team at NTNU \cite{sjalander+:2019epic} for providing essential computational resources, and to Schibsted and the National Library of Norway (Nasjonalbiblioteket) for supplying the crucial dataset for our research.

\newpage

\appendix
\section{Appendix}
The appendix contains instructions on how to access the dataset, all prompt templates used in our paper, supplementary results for PersonalSum and Topic-centric PersonalSum, an error case analysis of PersonalSum, and the statistics for Topic-centric PersonalSum.

\subsection{Accessibility}
All the resources in our work are accessible online \footnote{\url{https://github.com/SmartmediaAI/PersonalSum}, and \url{https://huggingface.co/datasets/PersonalLab/PersonalSum} with a Croissant metadata record.}, including dataset card and related code for data collection and quality control on Amazon Turk. The licensing for the dataset is under a CC BY-NC 4.0\footnote{\url{https://creativecommons.org/licenses/by-nc/4.0/}}. We will consistently maintain and update the resources to ensure long-term usability.

\subsection{Prompts}
\begin{figure}[H]
\centering
\epsfig{figure=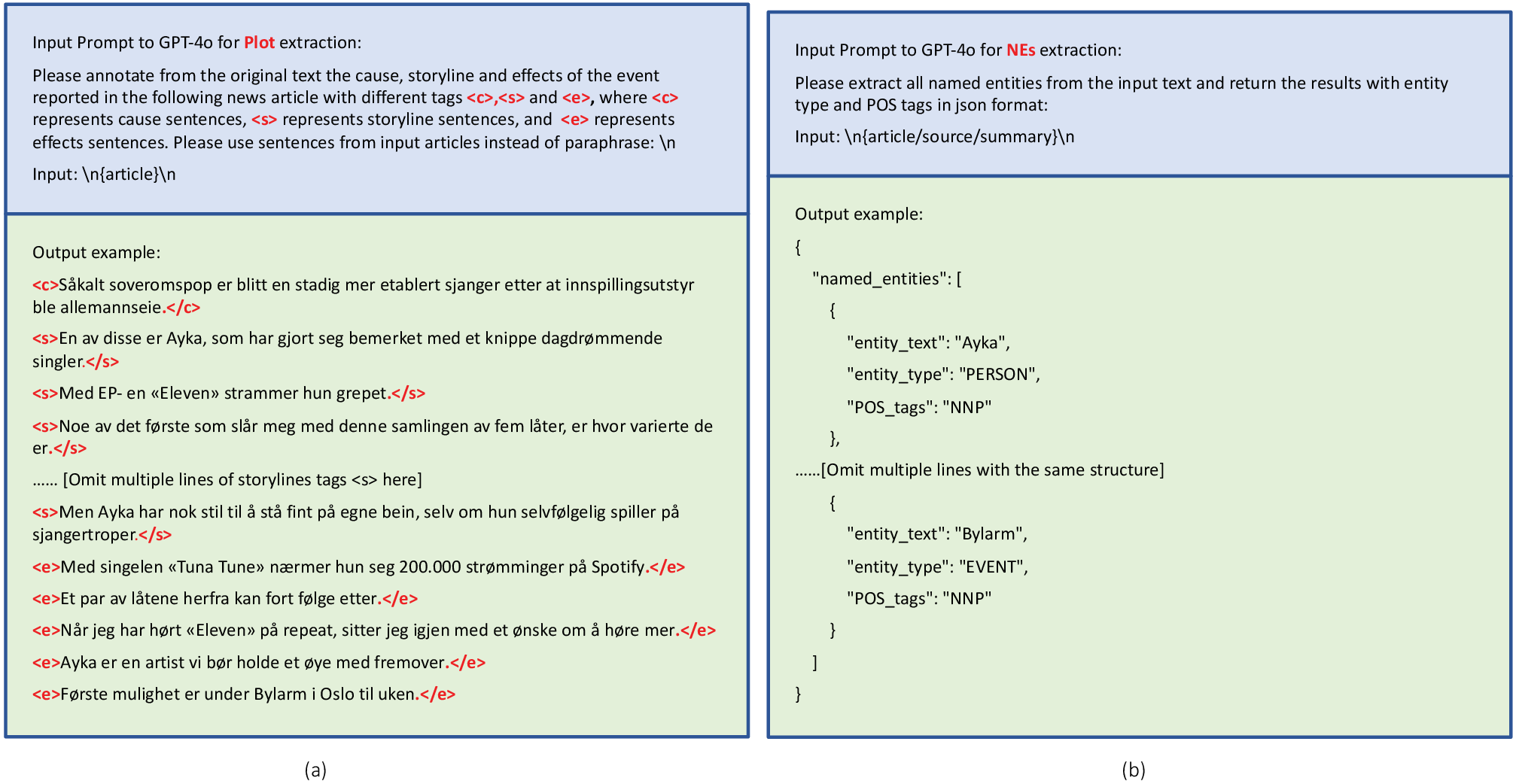, width=400pt}
\setlength{\abovecaptionskip}{2pt}
\setlength{\belowcaptionskip}{0pt}
\caption{Prompt template using GPT-4o on (a) Plot extraction, and (b) Named Entity (NE) recognition. } \label{fig:prompt_ne_plot}
\vspace{-7pt}
\end{figure}

\begin{figure}[H]
\centering
\epsfig{figure=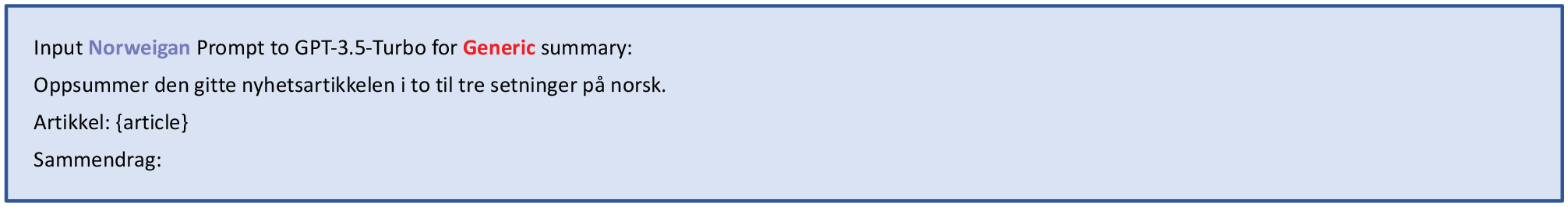, width=400pt}
\setlength{\abovecaptionskip}{2pt}
\setlength{\belowcaptionskip}{0pt}
\caption{Prompt for generic summary generation.} \label{fig:prompt_generic}
\end{figure}

\begin{figure}[H]
\centering
\epsfig{figure=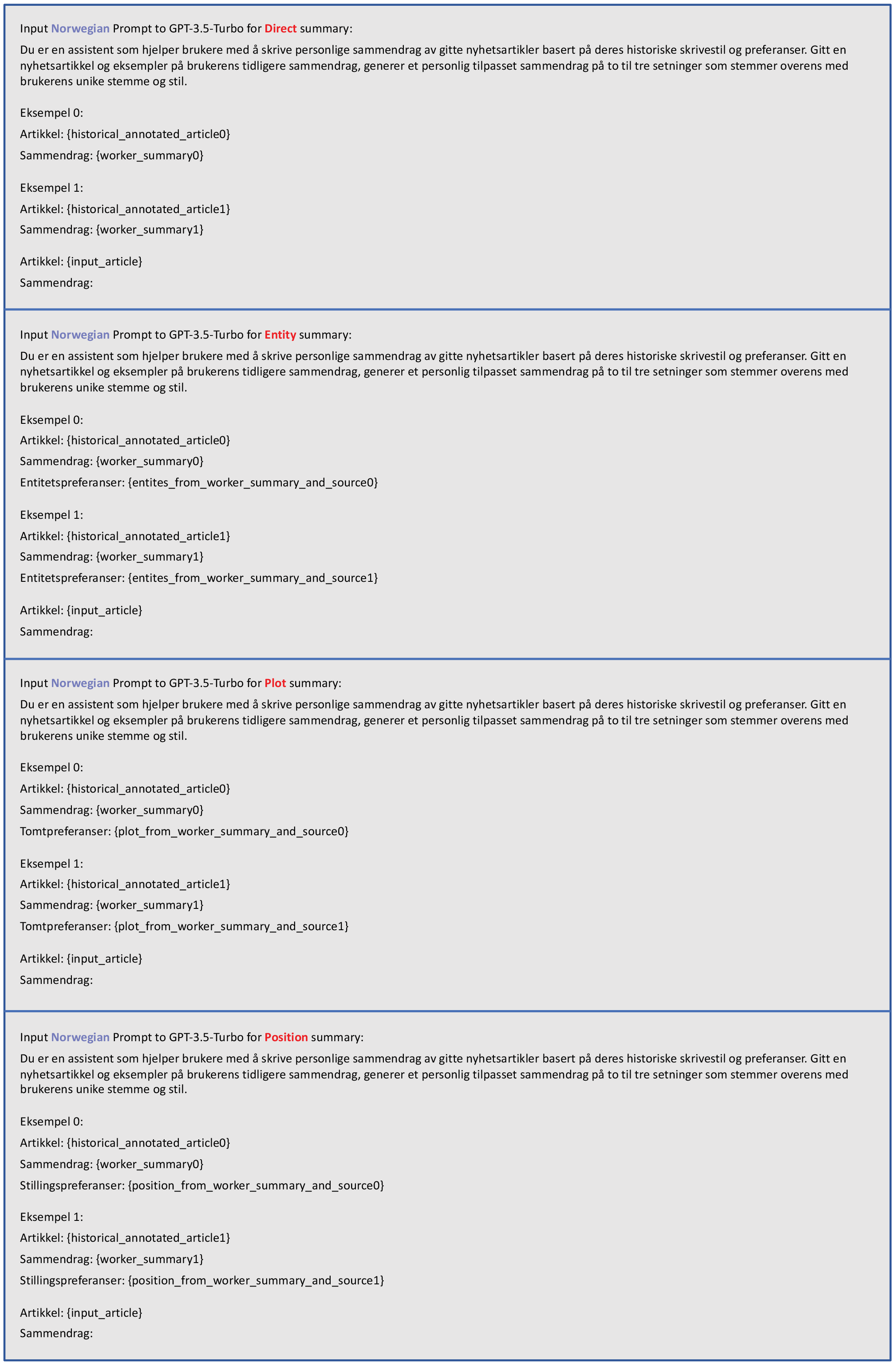, width=400pt}
\setlength{\abovecaptionskip}{2pt}
\setlength{\belowcaptionskip}{0pt}
\caption{Prompts for generating personalized summaries.} \label{fig:prompt_personalized}
\end{figure}

\begin{figure}[H]
\centering
\epsfig{figure=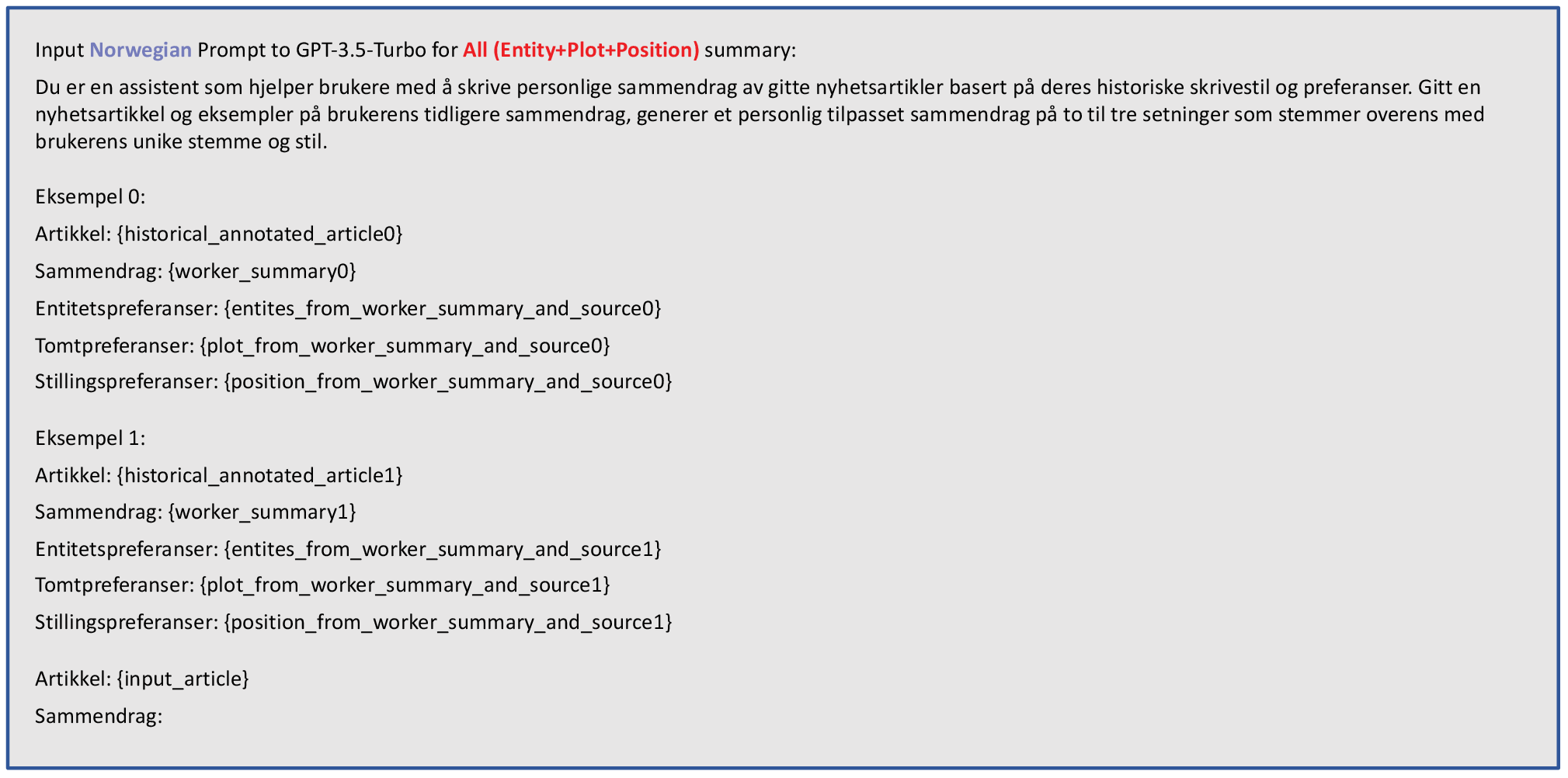, width=400pt}
\setlength{\abovecaptionskip}{2pt}
\setlength{\belowcaptionskip}{0pt}
\caption{Prompt for generating personalized summaries with all factors.} \label{fig:prompt_all}
\end{figure}

\begin{figure}[H]
\centering
\epsfig{figure=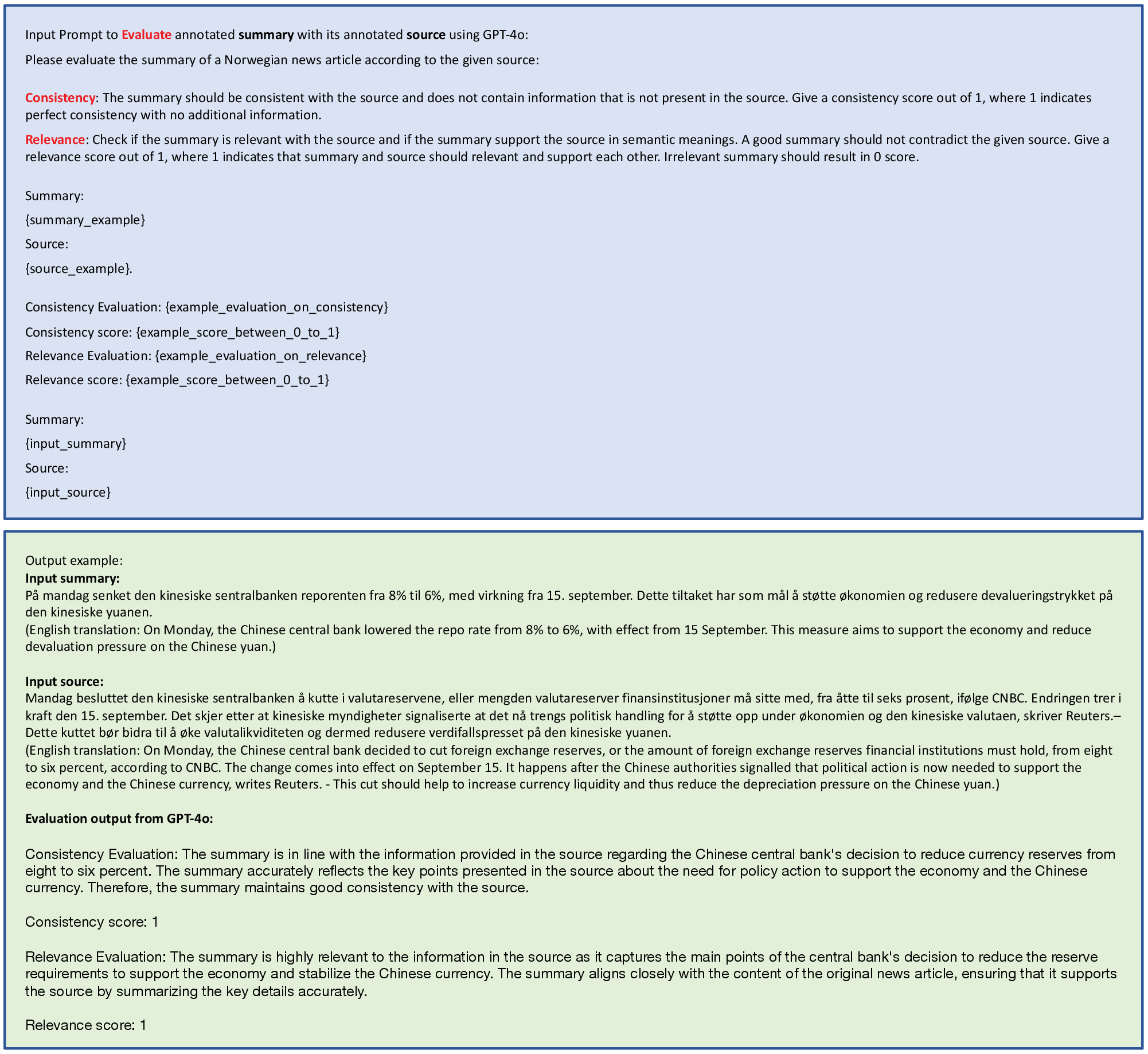, width=400pt}
\setlength{\abovecaptionskip}{2pt}
\setlength{\belowcaptionskip}{0pt}
\caption{Prompt for evaluating annotated summary w.r.t. its source using GPT-4o and an example output.} \label{fig:evaluation_prompt}
\end{figure}

This section lists all prompts used in the paper. Figure \ref{fig:prompt_ne_plot}(a) and (b) show the prompts used to extract plot and named entities (NEs) from the input article using GPT-4o, respectively. Figure \ref{fig:prompt_generic} and Figure \ref{fig:prompt_personalized} respectively show the Norwegian prompts we adopted to generate generic summaries (Generic), personalized summaries with only user history (Direct), personalized summaries with user history and entity (Entity)/plot (Plot)/position (Position) factor for all models in our experiments. When testing different combinations of factors, we add the associated factors to the user's history. For example, when exploring all combinations of factors (All), the prompt is as shown in Figure \ref{fig:prompt_all}. Figure \ref{fig:evaluation_prompt} shows the prompt used to control the quality of annotated summary in Section 3.1.

\subsection{Supplementary results for PersonalSum}

\begin{table}[H]
  \scriptsize
  \caption{\textcolor{red}{5-shot} experimental results of different LLMs on \textcolor{red}{PersonalSum}. Best results are on bold and the second best results are underlined.}
  \label{5-shot-results}
  \centering
  \setlength\tabcolsep{3.9pt}
  \begin{tabular}{p{0.9cm}<{\centering}|p{1.1cm}|p{0.8cm}<{\centering}p{0.8cm}<{\centering}p{0.8cm}<{\centering}p{0.8cm}<{\centering}p{0.8cm}<{\centering}p{1.1cm}<{\centering}p{1.45cm}<{\centering}p{1.25cm}<{\centering}p{0.8cm}<{\centering}}
    \toprule
     Models  & Metrics & Generic & Direct & Entity & Plot & Position & Entity+Plot & Entity+Position & Plot+Position & All \\ \midrule
    \multirow{8}{=}{\centering GPT-3.5 Turbo} & Rouge-1  & \tabincell{c}{37.53\\{\tiny $\pm$14.74}}  & \tabincell{c}{\textbf{38.10}\\{\tiny $\pm$14.88}} & \tabincell{c}{37.62\\{\tiny $\pm$15.19}} & \tabincell{c}{37.32\\{\tiny $\pm$16.04}} & \tabincell{c}{\underline{37.73}\\{\tiny $\pm$15.73}} & \tabincell{c}{37.32\\{\tiny $\pm$15.28}} & \tabincell{c}{37.62\\{\tiny $\pm$15.45}} & \tabincell{c}{37.23\\{\tiny $\pm$15.76}} & \tabincell{c}{37.66\\{\tiny $\pm$15.44}}\\
   \cmidrule(r){2-11}  & Rouge-2 & \tabincell{c}{16.83\\{\tiny $\pm$12.96}}  & \tabincell{c}{\textbf{17.40}\\{\tiny $\pm$13.11}} & \tabincell{c}{17.10\\{\tiny $\pm$13.10}} & \tabincell{c}{17.11\\{\tiny $\pm$13.56}} & \tabincell{c}{\underline{17.29}\\{\tiny $\pm$13.61}} & \tabincell{c}{16.76\\{\tiny $\pm$13.25}} & \tabincell{c}{17.04\\{\tiny $\pm$13.42}} & \tabincell{c}{16.83\\{\tiny $\pm$13.51}} & \tabincell{c}{17.22\\{\tiny $\pm$13.47}}\\
    \cmidrule(r){2-11} & Rouge-L  & \tabincell{c}{26.77\\{\tiny $\pm$13.26}}  & \tabincell{c}{\underline{27.08}\\{\tiny $\pm$13.27}} & \tabincell{c}{26.70\\{\tiny $\pm$13.48}} & \tabincell{c}{26.82\\{\tiny $\pm$14.06}} & \tabincell{c}{\textbf{27.10}\\{\tiny $\pm$13.95}} & \tabincell{c}{26.47\\{\tiny $\pm$13.55}} & \tabincell{c}{26.79\\{\tiny $\pm$13.62}} & \tabincell{c}{26.64\\{\tiny $\pm$13.97}} & \tabincell{c}{26.94\\{\tiny $\pm$13.66}}\\
    \cmidrule(r){2-11} & BERTScore  & \tabincell{c}{74.95\\{\tiny $\pm$5.35}}  & \tabincell{c}{\textbf{75.10}\\{\tiny $\pm$5.32}} & \tabincell{c}{74.79\\{\tiny $\pm$5.50}} & \tabincell{c}{74.71\\{\tiny $\pm$6.07}} & \tabincell{c}{\underline{74.96}\\{\tiny $\pm$5.92}} & \tabincell{c}{74.71\\{\tiny $\pm$5.67}} & \tabincell{c}{74.88\\{\tiny $\pm$5.66}} & \tabincell{c}{74.80\\{\tiny $\pm$5.87}} & \tabincell{c}{74.87\\{\tiny $\pm$5.73}}\\ \midrule[0.3pt]
    \multirow{8}{=}{\centering Gemini 1.0 Pro} & Rouge-1  & \tabincell{c}{32.27\\{\tiny $\pm$13.32}}  & \tabincell{c}{33.50\\{\tiny $\pm$13.63}} & \tabincell{c}{33.74\\{\tiny $\pm$13.64}} & \tabincell{c}{32.72\\{\tiny $\pm$13.48}} & \tabincell{c}{\textbf{33.91}\\{\tiny $\pm$13.65}} & \tabincell{c}{\underline{33.89}\\{\tiny $\pm$13.14}} & \tabincell{c}{33.85\\{\tiny $\pm$13.56}} & \tabincell{c}{33.66\\{\tiny $\pm$13.59}} & \tabincell{c}{33.64\\{\tiny $\pm$13.17}}\\
   \cmidrule(r){2-11}  & Rouge-2 & \tabincell{c}{12.35\\{\tiny $\pm$10.65}}  & \tabincell{c}{13.18\\{\tiny $\pm$11.08}} & \tabincell{c}{13.29\\{\tiny $\pm$10.73}} & \tabincell{c}{12.51\\{\tiny $\pm$10.83}} & \tabincell{c}{13.46\\{\tiny $\pm$11.57}} & \tabincell{c}{13.06\\{\tiny $\pm$10.84}} & \tabincell{c}{\textbf{13.61}\\{\tiny $\pm$11.03}} & \tabincell{c}{\underline{13.54}\\{\tiny $\pm$11.38}} & \tabincell{c}{13.14\\{\tiny $\pm$10.64}}\\
   \cmidrule(r){2-11}  & Rouge-L & \tabincell{c}{22.94\\{\tiny $\pm$11.62}}  & \tabincell{c}{23.89\\{\tiny $\pm$11.99}} & \tabincell{c}{23.96\\{\tiny $\pm$11.92}} & \tabincell{c}{23.38\\{\tiny $\pm$11.87}} & \tabincell{c}{\textbf{24.36}\\{\tiny $\pm$12.09}} & \tabincell{c}{23.99\\{\tiny $\pm$11.77}} & \tabincell{c}{23.93\\{\tiny $\pm$11.98}} & \tabincell{c}{\underline{24.10}\\{\tiny $\pm$12.10}} & \tabincell{c}{23.95\\{\tiny $\pm$11.55}}\\
   \cmidrule(r){2-11} & BERTScore  & \tabincell{c}{73.53\\{\tiny $\pm$5.14}}  & \tabincell{c}{73.95\\{\tiny $\pm$5.30}} & \tabincell{c}{73.94\\{\tiny $\pm$5.15}} & \tabincell{c}{73.64\\{\tiny $\pm$5.00}} & \tabincell{c}{\textbf{73.98}\\{\tiny $\pm$5.27}} & \tabincell{c}{73.90\\{\tiny $\pm$5.06}} & \tabincell{c}{73.88\\{\tiny $\pm$5.10}} & \tabincell{c}{\underline{73.95}\\{\tiny $\pm$5.13}} & \tabincell{c}{73.93\\{\tiny $\pm$5.13}}\\ \midrule[0.3pt]
    \multirow{8}{=}{\centering NorwAI-Mixtral-8x7B-instruct} & Rouge-1  & \tabincell{c}{34.16\\{\tiny $\pm$12.99}}  & \tabincell{c}{\textbf{34.32}\\{\tiny $\pm$13.56}} & \tabincell{c}{33.79\\{\tiny $\pm$13.67}} & \tabincell{c}{\underline{34.21}\\{\tiny $\pm$13.63}} & \tabincell{c}{33.93\\{\tiny $\pm$13.80}} & \tabincell{c}{33.96\\{\tiny $\pm$13.99}} & \tabincell{c}{34.08\\{\tiny $\pm$13.85}} & \tabincell{c}{34.15\\{\tiny $\pm$13.88}} & \tabincell{c}{33.75\\{\tiny $\pm$13.99}}\\
   \cmidrule(r){2-11}  & Rouge-2 & \tabincell{c}{13.42\\{\tiny $\pm$10.50}}  & \tabincell{c}{13.86\\{\tiny $\pm$10.98}} & \tabincell{c}{13.66\\{\tiny $\pm$11.11}} & \tabincell{c}{13.87\\{\tiny $\pm$11.13}} & \tabincell{c}{13.81\\{\tiny $\pm$11.17}} & \tabincell{c}{\textbf{13.98}\\{\tiny $\pm$11.30}} & \tabincell{c}{13.73\\{\tiny $\pm$11.21}} & \tabincell{c}{13.78\\{\tiny $\pm$11.21}} & \tabincell{c}{\underline{13.90}\\{\tiny $\pm$11.38}}\\
   \cmidrule(r){2-11}  & Rouge-L & \tabincell{c}{23.64\\{\tiny $\pm$10.81}}  & \tabincell{c}{\textbf{24.33}\\{\tiny $\pm$11.47}} & \tabincell{c}{24.11\\{\tiny $\pm$11.77}} & \tabincell{c}{\underline{24.33}\\{\tiny $\pm$11.67}} & \tabincell{c}{24.09\\{\tiny $\pm$11.57}} & \tabincell{c}{24.27\\{\tiny $\pm$11.94}} & \tabincell{c}{24.29\\{\tiny $\pm$11.80}} & \tabincell{c}{24.26\\{\tiny $\pm$11.78}} & \tabincell{c}{24.21\\{\tiny $\pm$11.93}}\\
   \cmidrule(r){2-11} & BERTScore  & \tabincell{c}{73.51\\{\tiny $\pm$4.70}}  & \tabincell{c}{\underline{73.90}\\{\tiny $\pm$4.97}} & \tabincell{c}{73.86\\{\tiny $\pm$5.05}} & \tabincell{c}{73.90\\{\tiny $\pm$5.04}} & \tabincell{c}{73.90\\{\tiny $\pm$5.06}} & \tabincell{c}{73.89\\{\tiny $\pm$5.14}} & \tabincell{c}{\textbf{73.93}\\{\tiny $\pm$5.03}} & \tabincell{c}{73.86\\{\tiny $\pm$5.12}} & \tabincell{c}{73.89\\{\tiny $\pm$5.15}}\\
    \bottomrule
  \end{tabular}
\end{table}

\begin{table}[H]
  \scriptsize
  \caption{\textcolor{red}{10-shot} experimental results of different LLMs on \textcolor{red}{PersonalSum}. Best results are on bold and the second best results are underlined.}
  \label{5-shot-results}
  \centering
  \setlength\tabcolsep{3.9pt}
  \begin{tabular}{p{0.9cm}<{\centering}|p{1.1cm}|p{0.8cm}<{\centering}p{0.8cm}<{\centering}p{0.8cm}<{\centering}p{0.8cm}<{\centering}p{0.8cm}<{\centering}p{1.1cm}<{\centering}p{1.45cm}<{\centering}p{1.25cm}<{\centering}p{0.8cm}<{\centering}}
    \toprule
     Models  & Metrics & Generic & Direct & Entity & Plot & Position & Entity+Plot & Entity+Position & Plot+Position & All \\ \midrule
    \multirow{8}{=}{\centering GPT-3.5 Turbo} & Rouge-1  & \tabincell{c}{37.40\\{\tiny $\pm$15.28}}  & \tabincell{c}{\underline{37.41}\\{\tiny $\pm$15.16}} & \tabincell{c}{37.00\\{\tiny $\pm$15.47}} & \tabincell{c}{36.63\\{\tiny $\pm$16.22}} & \tabincell{c}{36.62\\{\tiny $\pm$16.01}} & \tabincell{c}{37.33\\{\tiny $\pm$15.68}} & \tabincell{c}{37.36\\{\tiny $\pm$15.79}} & \tabincell{c}{\textbf{37.64}\\{\tiny $\pm$15.76}} & \tabincell{c}{37.10\\{\tiny $\pm$15.97}}\\
   \cmidrule(r){2-11}  & Rouge-2 & \tabincell{c}{17.00\\{\tiny $\pm$13.45}}  & \tabincell{c}{16.95\\{\tiny $\pm$13.20}} & \tabincell{c}{17.02\\{\tiny $\pm$13.31}} & \tabincell{c}{16.78\\{\tiny $\pm$13.54}} & \tabincell{c}{16.62\\{\tiny $\pm$13.60}} & \tabincell{c}{16.99\\{\tiny $\pm$13.13}} & \tabincell{c}{\underline{17.29}\\{\tiny $\pm$13.70}} & \tabincell{c}{\textbf{17.34}\\{\tiny $\pm$13.69}} & \tabincell{c}{17.01\\{\tiny $\pm$13.36}}\\
    \cmidrule(r){2-11} & Rouge-L  & \tabincell{c}{\underline{26.94}\\{\tiny $\pm$13.69}}  & \tabincell{c}{26.59\\{\tiny $\pm$13.35}} & \tabincell{c}{26.43\\{\tiny $\pm$13.64}} & \tabincell{c}{26.14\\{\tiny $\pm$14.08}} & \tabincell{c}{26.22\\{\tiny $\pm$14.05}} & \tabincell{c}{26.64\\{\tiny $\pm$13.59}} & \tabincell{c}{26.87\\{\tiny $\pm$14.23}} & \tabincell{c}{\textbf{27.06}\\{\tiny $\pm$13.93}} & \tabincell{c}{26.60\\{\tiny $\pm$13.89}}\\
    \cmidrule(r){2-11} & BERTScore  & \tabincell{c}{\textbf{74.99}\\{\tiny $\pm$5.57}}  & \tabincell{c}{74.84\\{\tiny $\pm$5.38}} & \tabincell{c}{74.66\\{\tiny $\pm$5.61}} & \tabincell{c}{74.46\\{\tiny $\pm$5.95}} & \tabincell{c}{74.51\\{\tiny $\pm$6.05}} & \tabincell{c}{74.70\\{\tiny $\pm$5.72}} & \tabincell{c}{74.86\\{\tiny $\pm$5.75}} & \tabincell{c}{\underline{74.92}\\{\tiny $\pm$5.77}} & \tabincell{c}{74.66\\{\tiny $\pm$5.93}}\\ \midrule[0.3pt]
    \multirow{8}{=}{\centering Gemini 1.0 Pro} & Rouge-1  & \tabincell{c}{33.22\\{\tiny $\pm$13.22}}  & \tabincell{c}{34.79\\{\tiny $\pm$14.11}} & \tabincell{c}{34.17\\{\tiny $\pm$13.79}} & \tabincell{c}{\textbf{35.78}\\{\tiny $\pm$14.21}} & \tabincell{c}{34.82\\{\tiny $\pm$14.11}} & \tabincell{c}{\underline{35.01}\\{\tiny $\pm$13.78}} & \tabincell{c}{34.46\\{\tiny $\pm$14.26}} & \tabincell{c}{34.70\\{\tiny $\pm$14.56}} & \tabincell{c}{34.63\\{\tiny $\pm$14.37}}\\
   \cmidrule(r){2-11}  & Rouge-2 & \tabincell{c}{12.82\\{\tiny $\pm$10.22}}  & \tabincell{c}{14.39\\{\tiny $\pm$11.16}} & \tabincell{c}{13.56\\{\tiny $\pm$11.41}} & \tabincell{c}{\textbf{14.55}\\{\tiny $\pm$11.81}} & \tabincell{c}{14.05\\{\tiny $\pm$11.34}} & \tabincell{c}{\underline{14.43}\\{\tiny $\pm$11.12}} & \tabincell{c}{13.86\\{\tiny $\pm$11.47}} & \tabincell{c}{14.25\\{\tiny $\pm$11.65}} & \tabincell{c}{13.86\\{\tiny $\pm$11.36}}\\
   \cmidrule(r){2-11}  & Rouge-L & \tabincell{c}{23.06\\{\tiny $\pm$11.22}}  & \tabincell{c}{\underline{24.65}\\{\tiny $\pm$11.82}} & \tabincell{c}{23.83\\{\tiny $\pm$11.90}} & \tabincell{c}{\textbf{24.93}\\{\tiny $\pm$11.84}} & \tabincell{c}{23.95\\{\tiny $\pm$11.51}} & \tabincell{c}{24.43\\{\tiny $\pm$11.61}} & \tabincell{c}{23.70\\{\tiny $\pm$11.68}} & \tabincell{c}{24.38\\{\tiny $\pm$12.06}} & \tabincell{c}{24.05\\{\tiny $\pm$11.60}}\\
   \cmidrule(r){2-11} & BERTScore  & \tabincell{c}{73.97\\{\tiny $\pm$4.62}}  & \tabincell{c}{74.26\\{\tiny $\pm$4.74}} & \tabincell{c}{74.00\\{\tiny $\pm$4.79}} & \tabincell{c}{\textbf{74.56}\\{\tiny $\pm$4.63}} & \tabincell{c}{74.33\\{\tiny $\pm$4.69}} & \tabincell{c}{\underline{74.37}\\{\tiny $\pm$4.79}} & \tabincell{c}{74.04\\{\tiny $\pm$4.99}} & \tabincell{c}{74.25\\{\tiny $\pm$4.92}} & \tabincell{c}{74.21\\{\tiny $\pm$4.84}}\\ \midrule[0.3pt]
    \multirow{8}{=}{\centering NorwAI-Mixtral-8x7B-instruct} & Rouge-1  & \tabincell{c}{32.97\\{\tiny $\pm$12.09}}  & \tabincell{c}{33.12\\{\tiny $\pm$11.89}} & \tabincell{c}{33.32\\{\tiny $\pm$12.45}} & \tabincell{c}{33.67\\{\tiny $\pm$12.36}} & \tabincell{c}{33.21\\{\tiny $\pm$12.80}} & \tabincell{c}{33.46\\{\tiny $\pm$12.81}} & \tabincell{c}{33.18\\{\tiny $\pm$12.25}} & \tabincell{c}{\textbf{34.05}\\{\tiny $\pm$12.25}} & \tabincell{c}{\underline{33.75}\\{\tiny $\pm$12.79}}\\
   \cmidrule(r){2-11}  & Rouge-2 & \tabincell{c}{12.67\\{\tiny $\pm$11.30}}  & \tabincell{c}{13.07\\{\tiny $\pm$10.89}} & \tabincell{c}{13.22\\{\tiny $\pm$11.10}} & \tabincell{c}{13.46\\{\tiny $\pm$11.03}} & \tabincell{c}{13.11\\{\tiny $\pm$11.17}} & \tabincell{c}{13.38\\{\tiny $\pm$10.60}} & \tabincell{c}{13.29\\{\tiny $\pm$10.36}} & \tabincell{c}{\textbf{13.67}\\{\tiny $\pm$10.71}} & \tabincell{c}{\underline{13.60}\\{\tiny $\pm$10.35}}\\
   \cmidrule(r){2-11}  & Rouge-L & \tabincell{c}{22.75\\{\tiny $\pm$10.71}}  & \tabincell{c}{23.03\\{\tiny $\pm$10.45}} & \tabincell{c}{23.16\\{\tiny $\pm$10.96}} & \tabincell{c}{23.52\\{\tiny $\pm$11.10}} & \tabincell{c}{23.39\\{\tiny $\pm$10.76}} & \tabincell{c}{23.46\\{\tiny $\pm$11.24}} & \tabincell{c}{23.37\\{\tiny $\pm$11.30}} & \tabincell{c}{\textbf{23.94}\\{\tiny $\pm$10.72}} & \tabincell{c}{\underline{23.83}\\{\tiny $\pm$11.71}}\\
   \cmidrule(r){2-11} & BERTScore  & \tabincell{c}{73.26\\{\tiny $\pm$4.56}}  & \tabincell{c}{73.39\\{\tiny $\pm$4.49}} & \tabincell{c}{73.51\\{\tiny $\pm$5.03}} & \tabincell{c}{\underline{73.78}\\{\tiny $\pm$4.86}} & \tabincell{c}{73.62\\{\tiny $\pm$5.13}} & \tabincell{c}{73.69\\{\tiny $\pm$4.74}} & \tabincell{c}{73.57\\{\tiny $\pm$5.01}} & \tabincell{c}{\textbf{73.84}\\{\tiny $\pm$4.82}} & \tabincell{c}{73.59\\{\tiny $\pm$5.20}}\\
    \bottomrule
  \end{tabular}
\end{table}

\subsection{Supplementary results for Topic-centric PersonalSum}

\begin{table}[H]
  \scriptsize
  \caption{\textcolor{red}{2-shot} experimental results of different LLMs on \textcolor{red}{Topic-centric PersonalSum}. Best results are on bold and the second best results are underlined.}
  \label{2-shot-results}
  \centering
  \setlength\tabcolsep{3.9pt}
  \begin{tabular}{p{0.9cm}<{\centering}|p{1.1cm}|p{0.8cm}<{\centering}p{0.8cm}<{\centering}p{0.8cm}<{\centering}p{0.8cm}<{\centering}p{0.8cm}<{\centering}p{1.1cm}<{\centering}p{1.45cm}<{\centering}p{1.25cm}<{\centering}p{0.8cm}<{\centering}}
    \toprule
     Models  & Metrics & Generic & Direct & Entity & Plot & Position & Entity+Plot & Entity+Position & Plot+Position & All \\ \midrule
    \multirow{8}{=}{\centering GPT-3.5 Turbo} & Rouge-1  & \tabincell{c}{37.14\\{\tiny $\pm$14.02}}  & \tabincell{c}{39.16\\{\tiny $\pm$14.13}} & \tabincell{c}{39.21\\{\tiny $\pm$14.36}} & \tabincell{c}{\underline{39.42}\\{\tiny $\pm$15.09}} & \tabincell{c}{39.37\\{\tiny $\pm$14.18}} & \tabincell{c}{38.78\\{\tiny $\pm$14.36}} & \tabincell{c}{\textbf{39.90}\\{\tiny $\pm$14.42}} & \tabincell{c}{39.30\\{\tiny $\pm$14.01}} & \tabincell{c}{39.07\\{\tiny $\pm$14.46}}\\
   \cmidrule(r){2-11}  & Rouge-2 & \tabincell{c}{16.43\\{\tiny $\pm$12.06}}  & \tabincell{c}{17.58\\{\tiny $\pm$12.30}} & \tabincell{c}{17.94\\{\tiny $\pm$12.78}} & \tabincell{c}{18.15\\{\tiny $\pm$12.84}} & \tabincell{c}{17.94\\{\tiny $\pm$12.42}} & \tabincell{c}{17.86\\{\tiny $\pm$12.43}} & \tabincell{c}{\textbf{18.81}\\{\tiny $\pm$12.75}} & \tabincell{c}{\underline{18.22}\\{\tiny $\pm$12.33}} & \tabincell{c}{17.61\\{\tiny $\pm$12.35}}\\
    \cmidrule(r){2-11} & Rouge-L  & \tabincell{c}{26.43\\{\tiny $\pm$12.30}}  & \tabincell{c}{27.23\\{\tiny $\pm$12.20}} & \tabincell{c}{27.72\\{\tiny $\pm$12.86}} & \tabincell{c}{28.03\\{\tiny $\pm$13.26}} & \tabincell{c}{27.70\\{\tiny $\pm$12.40}} & \tabincell{c}{27.27\\{\tiny $\pm$12.37}} & \tabincell{c}{\underline{28.06}\\{\tiny $\pm$12.56}} & \tabincell{c}{\textbf{28.12}\\{\tiny $\pm$12.67}} & \tabincell{c}{27.50\\{\tiny $\pm$12.54}}\\
    \cmidrule(r){2-11} & BERTScore  & \tabincell{c}{74.84\\{\tiny $\pm$5.13}}  & \tabincell{c}{75.57\\{\tiny $\pm$5.20}} & \tabincell{c}{75.60\\{\tiny $\pm$5.38}} & \tabincell{c}{75.59\\{\tiny $\pm$5.59}} & \tabincell{c}{75.66\\{\tiny $\pm$5.25}} & \tabincell{c}{75.51\\{\tiny $\pm$5.30}} & \tabincell{c}{\textbf{75.75}\\{\tiny $\pm$5.37}} & \tabincell{c}{\underline{75.73}\\{\tiny $\pm$5.13}} & \tabincell{c}{75.54\\{\tiny $\pm$5.32}}\\ \midrule[0.3pt]
    \multirow{8}{=}{\centering Gemini 1.0 Pro} & Rouge-1  & \tabincell{c}{35.46\\{\tiny $\pm$12.84}}  & \tabincell{c}{\textbf{36.48}\\{\tiny $\pm$12.96}} & \tabincell{c}{36.30\\{\tiny $\pm$13.25}} & \tabincell{c}{35.26\\{\tiny $\pm$14.32}} & \tabincell{c}{35.14\\{\tiny $\pm$14.53}} & \tabincell{c}{\underline{36.46}\\{\tiny $\pm$14.10}} & \tabincell{c}{36.32\\{\tiny $\pm$13.54}} & \tabincell{c}{36.18\\{\tiny $\pm$13.86}} & \tabincell{c}{36.33\\{\tiny $\pm$13.81}}\\
   \cmidrule(r){2-11}  & Rouge-2 & \tabincell{c}{14.52\\{\tiny $\pm$11.02}}  & \tabincell{c}{14.74\\{\tiny $\pm$10.41}} & \tabincell{c}{15.04\\{\tiny $\pm$10.35}} & \tabincell{c}{13.89\\{\tiny $\pm$10.46}} & \tabincell{c}{14.24\\{\tiny $\pm$10.93}} & \tabincell{c}{\textbf{15.19}\\{\tiny $\pm$11.26}} & \tabincell{c}{14.99\\{\tiny $\pm$11.38}} & \tabincell{c}{\underline{15.11}\\{\tiny $\pm$11.22}} & \tabincell{c}{15.06\\{\tiny $\pm$11.11}}\\
   \cmidrule(r){2-11}  & Rouge-L & \tabincell{c}{24.88\\{\tiny $\pm$11.16}}  & \tabincell{c}{25.74\\{\tiny $\pm$11.59}} & \tabincell{c}{25.73\\{\tiny $\pm$11.17}} & \tabincell{c}{24.54\\{\tiny $\pm$12.20}} & \tabincell{c}{24.63\\{\tiny $\pm$11.95}} & \tabincell{c}{\textbf{25.88}\\{\tiny $\pm$12.56}} & \tabincell{c}{25.75\\{\tiny $\pm$12.17}} & \tabincell{c}{25.59\\{\tiny $\pm$12.42}} & \tabincell{c}{\underline{25.76}\\{\tiny $\pm$11.89}}\\
   \cmidrule(r){2-11} & BERTScore  & \tabincell{c}{74.45\\{\tiny $\pm$4.95}}  & \tabincell{c}{\underline{74.82}\\{\tiny $\pm$4.98}} & \tabincell{c}{74.66\\{\tiny $\pm$5.20}} & \tabincell{c}{74.32\\{\tiny $\pm$5.61}} & \tabincell{c}{73.85\\{\tiny $\pm$6.41}} & \tabincell{c}{\textbf{74.85}\\{\tiny $\pm$5.49}} & \tabincell{c}{74.72\\{\tiny $\pm$5.44}} & \tabincell{c}{74.60\\{\tiny $\pm$5.54}} & \tabincell{c}{74.76\\{\tiny $\pm$5.11}}\\ \midrule[0.3pt]
    \multirow{8}{=}{\centering NorwAI-Mixtral-8x7B-instruct} & Rouge-1  & \tabincell{c}{32.80\\{\tiny $\pm$11.98}}  & \tabincell{c}{\underline{33.87}\\{\tiny $\pm$12.14}} & \tabincell{c}{33.20\\{\tiny $\pm$11.85}} & \tabincell{c}{33.23\\{\tiny $\pm$13.19}} & \tabincell{c}{33.65\\{\tiny $\pm$12.49}} & \tabincell{c}{33.19\\{\tiny $\pm$12.57}} & \tabincell{c}{\textbf{34.32}\\{\tiny $\pm$12.90}} & \tabincell{c}{33.58\\{\tiny $\pm$12.71}} & \tabincell{c}{33.34\\{\tiny $\pm$13.17}}\\
   \cmidrule(r){2-11}  & Rouge-2 & \tabincell{c}{12.29\\{\tiny $\pm$9.47}}  & \tabincell{c}{13.28\\{\tiny $\pm$9.55}} & \tabincell{c}{12.61\\{\tiny $\pm$8.87}} & \tabincell{c}{13.07\\{\tiny $\pm$9.92}} & \tabincell{c}{13.04\\{\tiny $\pm$9.65}} & \tabincell{c}{12.92\\{\tiny $\pm$9.68}} & \tabincell{c}{\textbf{13.82}\\{\tiny $\pm$10.22}} & \tabincell{c}{12.66\\{\tiny $\pm$9.31}} & \tabincell{c}{\underline{13.35}\\{\tiny $\pm$10.21}}\\
   \cmidrule(r){2-11}  & Rouge-L & \tabincell{c}{22.66\\{\tiny $\pm$10.26}}  & \tabincell{c}{\underline{23.62}\\{\tiny $\pm$10.26}} & \tabincell{c}{22.68\\{\tiny $\pm$9.24}} & \tabincell{c}{23.01\\{\tiny $\pm$10.78}} & \tabincell{c}{23.45\\{\tiny $\pm$10.17}} & \tabincell{c}{22.96\\{\tiny $\pm$10.18}} & \tabincell{c}{\textbf{23.93}\\{\tiny $\pm$10.97}} & \tabincell{c}{23.23\\{\tiny $\pm$10.14}} & \tabincell{c}{23.32\\{\tiny $\pm$10.82}}\\
   \cmidrule(r){2-11} & BERTScore  & \tabincell{c}{73.15\\{\tiny $\pm$4.57}}  & \tabincell{c}{\underline{73.77}\\{\tiny $\pm$4.52}} & \tabincell{c}{73.63\\{\tiny $\pm$4.41}} & \tabincell{c}{73.73\\{\tiny $\pm$4.75}} & \tabincell{c}{73.60\\{\tiny $\pm$4.51}} & \tabincell{c}{73.69\\{\tiny $\pm$4.72}} & \tabincell{c}{\textbf{74.13}\\{\tiny $\pm$4.80}} & \tabincell{c}{73.60\\{\tiny $\pm$4.72}} & \tabincell{c}{73.72\\{\tiny $\pm$4.80}}\\ \midrule[0.3pt]
    \multirow{8}{=}{\centering Meta-Llama3-70B-Instruct} & Rouge-1  & \tabincell{c}{35.98\\{\tiny $\pm$13.02}}  & \tabincell{c}{17.86\\{\tiny $\pm$14.97}} & \tabincell{c}{16.73\\{\tiny $\pm$13.82}} & \tabincell{c}{17.88\\{\tiny $\pm$16.50}} & \tabincell{c}{18.30\\{\tiny $\pm$16.23}} & \tabincell{c}{17.86\\{\tiny $\pm$15.96}} & \tabincell{c}{16.60\\{\tiny $\pm$14.99}} & \tabincell{c}{18.61\\{\tiny $\pm$16.45}} & \tabincell{c}{18.05\\{\tiny $\pm$15.80}}\\
   \cmidrule(r){2-11}  & Rouge-2 & \tabincell{c}{14.15\\{\tiny $\pm$10.29}}  & \tabincell{c}{7.20\\{\tiny $\pm$8.88}} & \tabincell{c}{6.74\\{\tiny $\pm$8.00}} & \tabincell{c}{7.33\\{\tiny $\pm$9.36}} & \tabincell{c}{7.54\\{\tiny $\pm$9.61}} & \tabincell{c}{7.43\\{\tiny $\pm$8.98}} & \tabincell{c}{6.48\\{\tiny $\pm$8.24}} & \tabincell{c}{7.66\\{\tiny $\pm$9.41}} & \tabincell{c}{7.29\\{\tiny $\pm$9.30}}\\
   \cmidrule(r){2-11}  & Rouge-L & \tabincell{c}{24.25\\{\tiny $\pm$10.21}}  & \tabincell{c}{13.56\\{\tiny $\pm$10.86}} & \tabincell{c}{12.87\\{\tiny $\pm$9.97}} & \tabincell{c}{13.17\\{\tiny $\pm$11.67}} & \tabincell{c}{13.49\\{\tiny $\pm$11.60}} & \tabincell{c}{13.43\\{\tiny $\pm$11.35}} & \tabincell{c}{12.52\\{\tiny $\pm$10.95}} & \tabincell{c}{13.69\\{\tiny $\pm$11.63}} & \tabincell{c}{13.33\\{\tiny $\pm$11.28}}\\
   \cmidrule(r){2-11} & BERTScore  & \tabincell{c}{74.10\\{\tiny $\pm$4.99}}  & \tabincell{c}{71.03\\{\tiny $\pm$4.62}} & \tabincell{c}{69.93\\{\tiny $\pm$5.09}} & \tabincell{c}{70.05\\{\tiny $\pm$6.50}} & \tabincell{c}{70.48\\{\tiny $\pm$5.75}} & \tabincell{c}{70.31\\{\tiny $\pm$6.11}} & \tabincell{c}{69.78\\{\tiny $\pm$6.10}} & \tabincell{c}{70.59\\{\tiny $\pm$5.64}} & \tabincell{c}{70.03\\{\tiny $\pm$6.10}}\\
    \bottomrule
  \end{tabular}
\end{table}

\begin{table}
  \scriptsize
  \caption{\textcolor{red}{10-shot} experimental results of different LLMs on \textcolor{red}{Topic-centric PersonalSum}. Best results are on bold and the second best results are underlined.}
  \label{10-shot-results}
  \centering
  \setlength\tabcolsep{3.9pt}
  \begin{tabular}{p{0.9cm}<{\centering}|p{1.1cm}|p{0.8cm}<{\centering}p{0.8cm}<{\centering}p{0.8cm}<{\centering}p{0.8cm}<{\centering}p{0.8cm}<{\centering}p{1.1cm}<{\centering}p{1.45cm}<{\centering}p{1.25cm}<{\centering}p{0.8cm}<{\centering}}
    \toprule
     Models  & Metrics & Generic & Direct & Entity & Plot & Position & Entity+Plot & Entity+Position & Plot+Position & All \\ \midrule
    \multirow{8}{=}{\centering GPT-3.5 Turbo} & Rouge-1  & \tabincell{c}{37.36\\{\tiny $\pm$14.06}}  & \tabincell{c}{37.74\\{\tiny $\pm$14.30}} & \tabincell{c}{37.99\\{\tiny $\pm$13.87}} & \tabincell{c}{37.96\\{\tiny $\pm$15.51}} & \tabincell{c}{38.39\\{\tiny $\pm$13.85}} & \tabincell{c}{37.53\\{\tiny $\pm$15.69}} & \tabincell{c}{36.80\\{\tiny $\pm$15.29}} & \tabincell{c}{\textbf{38.89}\\{\tiny $\pm$13.94}} & \tabincell{c}{\underline{38.68}\\{\tiny $\pm$14.23}}\\
   \cmidrule(r){2-11}  & Rouge-2 & \tabincell{c}{16.70\\{\tiny $\pm$11.91}}  & \tabincell{c}{17.25\\{\tiny $\pm$12.46}} & \tabincell{c}{17.65\\{\tiny $\pm$11.69}} & \tabincell{c}{17.68\\{\tiny $\pm$12.37}} & \tabincell{c}{17.58\\{\tiny $\pm$11.60}} & \tabincell{c}{17.57\\{\tiny $\pm$12.83}} & \tabincell{c}{17.71\\{\tiny $\pm$12.77}} & \tabincell{c}{\textbf{18.01}\\{\tiny $\pm$11.58}} & \tabincell{c}{\underline{17.97}\\{\tiny $\pm$12.27}}\\
    \cmidrule(r){2-11} & Rouge-L  & \tabincell{c}{26.67\\{\tiny $\pm$12.26}}  & \tabincell{c}{27.19\\{\tiny $\pm$13.10}} & \tabincell{c}{27.09\\{\tiny $\pm$12.25}} & \tabincell{c}{26.72\\{\tiny $\pm$12.86}} & \tabincell{c}{27.18\\{\tiny $\pm$12.16}} & \tabincell{c}{26.89\\{\tiny $\pm$13.30}} & \tabincell{c}{27.48\\{\tiny $\pm$13.85}} & \tabincell{c}{\underline{27.63}\\{\tiny $\pm$11.86}} & \tabincell{c}{\textbf{28.07}\\{\tiny $\pm$12.60}}\\
    \cmidrule(r){2-11} & BERTScore  & \tabincell{c}{74.90\\{\tiny $\pm$5.12}}  & \tabincell{c}{75.17\\{\tiny $\pm$5.26}} & \tabincell{c}{75.09\\{\tiny $\pm$5.04}} & \tabincell{c}{74.93\\{\tiny $\pm$5.64}} & \tabincell{c}{75.16\\{\tiny $\pm$5.42}} & \tabincell{c}{75.31\\{\tiny $\pm$5.30}} & \tabincell{c}{75.20\\{\tiny $\pm$5.25}} & \tabincell{c}{\underline{75.50}\\{\tiny $\pm$4.91}} & \tabincell{c}{\textbf{75.57}\\{\tiny $\pm$5.02}}\\ \midrule[0.3pt]
    \multirow{8}{=}{\centering Gemini 1.0 Pro} & Rouge-1  & \tabincell{c}{35.94\\{\tiny $\pm$12.65}}  & \tabincell{c}{37.45\\{\tiny $\pm$14.01}} & \tabincell{c}{\textbf{38.44}\\{\tiny $\pm$13.10}} & \tabincell{c}{37.08\\{\tiny $\pm$14.43}} & \tabincell{c}{37.52\\{\tiny $\pm$13.50}} & \tabincell{c}{37.32\\{\tiny $\pm$13.88}} & \tabincell{c}{36.89\\{\tiny $\pm$13.53}} & \tabincell{c}{37.14\\{\tiny $\pm$14.27}} & \tabincell{c}{\underline{37.90}\\{\tiny $\pm$13.51}}\\
   \cmidrule(r){2-11}  & Rouge-2 & \tabincell{c}{14.60\\{\tiny $\pm$10.45}}  & \tabincell{c}{16.05\\{\tiny $\pm$11.57}} & \tabincell{c}{\underline{16.33}\\{\tiny $\pm$10.77}} & \tabincell{c}{15.52\\{\tiny $\pm$11.36}} & \tabincell{c}{16.16\\{\tiny $\pm$11.27}} & \tabincell{c}{16.20\\{\tiny $\pm$11.33}} & \tabincell{c}{15.76\\{\tiny $\pm$10.93}} & \tabincell{c}{\textbf{16.51}\\{\tiny $\pm$12.30}} & \tabincell{c}{16.24\\{\tiny $\pm$11.37}}\\
   \cmidrule(r){2-11}  & Rouge-L & \tabincell{c}{24.87\\{\tiny $\pm$10.85}}  & \tabincell{c}{\underline{26.76}\\{\tiny $\pm$12.67}} & \tabincell{c}{26.55\\{\tiny $\pm$11.49}} & \tabincell{c}{25.85\\{\tiny $\pm$12.08}} & \tabincell{c}{26.24\\{\tiny $\pm$12.43}} & \tabincell{c}{26.27\\{\tiny $\pm$12.72}} & \tabincell{c}{26.12\\{\tiny $\pm$12.25}} & \tabincell{c}{\textbf{26.91}\\{\tiny $\pm$13.47}} & \tabincell{c}{26.40\\{\tiny $\pm$12.13}}\\
   \cmidrule(r){2-11} & BERTScore  & \tabincell{c}{74.74\\{\tiny $\pm$4.96}}  & \tabincell{c}{74.97\\{\tiny $\pm$5.20}} & \tabincell{c}{75.04\\{\tiny $\pm$4.99}} & \tabincell{c}{74.65\\{\tiny $\pm$5.63}} & \tabincell{c}{74.92\\{\tiny $\pm$5.19}} & \tabincell{c}{74.79\\{\tiny $\pm$5.52}} & \tabincell{c}{74.76\\{\tiny $\pm$5.16}} & \tabincell{c}{\underline{75.14}\\{\tiny $\pm$5.64}} & \tabincell{c}{\textbf{75.18}\\{\tiny $\pm$5.24}}\\ \midrule[0.3pt]
    \multirow{8}{=}{\centering NorwAI-Mixtral-8x7B-instruct} & Rouge-1  & \tabincell{c}{34.00\\{\tiny $\pm$12.94}}  & \tabincell{c}{\underline{35.78}\\{\tiny $\pm$13.36}} & \tabincell{c}{34.93\\{\tiny $\pm$13.58}} & \tabincell{c}{34.81\\{\tiny $\pm$13.53}} & \tabincell{c}{35.15\\{\tiny $\pm$13.05}} & \tabincell{c}{\textbf{36.00}\\{\tiny $\pm$13.94}} & \tabincell{c}{35.03\\{\tiny $\pm$13.90}} & \tabincell{c}{34.96\\{\tiny $\pm$13.48}} & \tabincell{c}{35.14\\{\tiny $\pm$13.64}}\\
   \cmidrule(r){2-11}  & Rouge-2 & \tabincell{c}{13.33\\{\tiny $\pm$9.96}}  & \tabincell{c}{\underline{15.13}\\{\tiny $\pm$9.80}} & \tabincell{c}{14.28\\{\tiny $\pm$10.30}} & \tabincell{c}{14.17\\{\tiny $\pm$10.21}} & \tabincell{c}{14.41\\{\tiny $\pm$9.76}} & \tabincell{c}{\textbf{15.50}\\{\tiny $\pm$10.54}} & \tabincell{c}{14.39\\{\tiny $\pm$10.96}} & \tabincell{c}{14.67\\{\tiny $\pm$10.19}} & \tabincell{c}{14.51\\{\tiny $\pm$10.43}}\\
   \cmidrule(r){2-11}  & Rouge-L & \tabincell{c}{23.93\\{\tiny $\pm$10.85}}  & \tabincell{c}{24.91\\{\tiny $\pm$10.57}} & \tabincell{c}{24.78\\{\tiny $\pm$11.14}} & \tabincell{c}{24.22\\{\tiny $\pm$10.93}} & \tabincell{c}{24.44\\{\tiny $\pm$10.42}} & \tabincell{c}{\textbf{25.38}\\{\tiny $\pm$11.46}} & \tabincell{c}{24.78\\{\tiny $\pm$11.72}} & \tabincell{c}{24.64\\{\tiny $\pm$11.13}} & \tabincell{c}{\underline{24.96}\\{\tiny $\pm$11.33}}\\
   \cmidrule(r){2-11} & BERTScore  & \tabincell{c}{73.63\\{\tiny $\pm$4.86}}  & \tabincell{c}{\underline{74.38}\\{\tiny $\pm$4.93}} & \tabincell{c}{74.11\\{\tiny $\pm$4.89}} & \tabincell{c}{74.19\\{\tiny $\pm$5.03}} & \tabincell{c}{74.30\\{\tiny $\pm$5.03}} & \tabincell{c}{\textbf{74.43}\\{\tiny $\pm$5.16}} & \tabincell{c}{74.29\\{\tiny $\pm$5.27}} & \tabincell{c}{74.24\\{\tiny $\pm$5.11}} & \tabincell{c}{74.23\\{\tiny $\pm$5.07}}\\
    \bottomrule
  \end{tabular}
\end{table}

\begin{table}
  \scriptsize
  \caption{Entailment scores of different LLMs on PersonalSum. Best results are on bold.}
  \label{entailment-results}
  \centering
  \begin{tabular}{p{0.8cm}<{\centering}|p{1.7cm}|m{0.6cm}<{\centering}p{0.6cm}<{\centering}p{0.6cm}<{\centering}p{0.6cm}<{\centering}p{0.6cm}<{\centering}p{0.86cm}<{\centering}p{1.25cm}<{\centering}p{1.05cm}<{\centering}p{0.6cm}<{\centering}}
    \toprule
       & Models & Generic & Direct & Entity & Plot & Position & Entity+Plot & Entity+Position & Plot+Position & All \\ 
    \midrule[0.3pt]
    \rowcolor{lightgray!50} \multicolumn{11}{c}{PersonalSum Dataset}\\ \midrule[0.3pt]
    \multirow{4}{=}{\centering 2-shot} & GPT-3.5 Turbo  & 94.91  & 94.22 & 95.10 & 94.81 & 94.03 & 94.61 & 94.81 & \textbf{95.69} & 93.24\\
   \cmidrule(r){2-11}  & Gemini 1.0 Pro & 90.04  & 90.26 & \textbf{91.56} & 90.15 & 86.47 & 90.80 & 88.64 & 87.77 & 89.07\\
    \cmidrule(r){2-11} & NorwAI-Mixtral  & 93.14  & 95.00 & 97.16 & 96.87 & 95.40 & \textbf{97.26} & 96.47 & 96.57 & 95.10\\ \midrule[0.3pt]
    \multirow{4}{=}{\centering 5-shot} & GPT-3.5 Turbo  & 95.17  & 94.07 & \textbf{96.05} & 94.95 & 95.17 & \textbf{96.05} & 94.07 & 94.95 & 94.95\\
   \cmidrule(r){2-11}  & Gemini 1.0 Pro & 90.80  & 90.68 & \textbf{91.02} & 88.96 & 90.26 & 88.64 & 90.15 & 89.61 & 89.61\\
    \cmidrule(r){2-11} & NorwAI-Mixtral  & 93.70  & 94.33 & 96.53 & 96.22 & 96.53 & 97.37 & 96.01 & \textbf{97.58} & 96.11\\ \midrule[0.3pt]
    \multirow{4}{=}{\centering 10-shot} & GPT-3.5 Turbo  & 94.39  & 95.01 & 95.51 & \textbf{96.01} & 94.89 & 94.01 & 95.76 & 95.76 & 95.14\\
   \cmidrule(r){2-11}  & Gemini 1.0 Pro & 88.16  & 92.52 & 92.64 & 91.77 & 91.90 & 93.52 & 91.65 & 92.39 & \textbf{93.76}\\
    \cmidrule(r){2-11} & NorwAI-Mixtral  & 94.28  & 95.08 & \textbf{97.83} & 96.00 & 95.64 & 96.80 & 97.48 & 96.22 & 97.14\\ \midrule[0.3pt]
    \rowcolor{lightgray!50} \multicolumn{11}{c}{Topic-centric PersonalSum Dataset}\\ \midrule[0.3pt]
    \multirow{4}{=}{\centering 2-shot} & GPT-3.5 Turbo  & 91.50  & 92.71 & 88.66 & 90.69 & 92.31 & 89.88 & 90.28 & \textbf{93.12} & 91.90\\
   \cmidrule(r){2-11}  & Gemini 1.0 Pro & \textbf{95.15}  & 92.07 & 91.19 & 92.51 & 89.43 & 92.07 & 92.95 & 87.22 & 92.51\\
    \cmidrule(r){2-11} & NorwAI-Mixtral  & 90.28  & 93.93 & 96.76 & 97.98 & 94.74 & 95.14 & 94.33 & \textbf{98.38} & 95.95\\ \midrule[0.3pt]
    \multirow{4}{=}{\centering 5-shot} & GPT-3.5 Turbo  & 91.94  & 93.84 & 92.89 & 91.00 & 91.00 & \textbf{95.73} & 93.35 & 91.47 & 92.42\\
   \cmidrule(r){2-11}  & Gemini 1.0 Pro & 93.84  & 92.89 & 91.47 & 93.36 & \textbf{94.79} & \textbf{94.79} & 93.84 & 92.89 & 90.05\\
    \cmidrule(r){2-11} & NorwAI-Mixtral  & 91.24  & 95.39 & 95.85 & 94.47 & 94.01 & \textbf{96.77} & 94.93 & 96.31 & 94.93\\ \midrule[0.3pt]
    \multirow{4}{=}{\centering 10-shot} & GPT-3.5 Turbo  & 91.30  & 95.03 & 93.17 & \textbf{96.27} & 93.79 & 92.55 & 94.41 & 95.03 & 93.79\\
   \cmidrule(r){2-11}  & Gemini 1.0 Pro & \textbf{94.70}  & 91.39 & 93.38 & \textbf{94.70} & 91.39 & 94.04 & 90.73 & 88.74 & 88.74\\
    \cmidrule(r){2-11} & NorwAI-Mixtral  & 93.60  & 90.12 & \textbf{97.09} & 93.60 & 94.77 & 95.35 & 96.51 & 94.19 & \textbf{97.09}\\
    \bottomrule
  \end{tabular}
\end{table}

\subsection{Error Case Analysis on PersonalSum}
\begin{figure}
\centering
\epsfig{figure=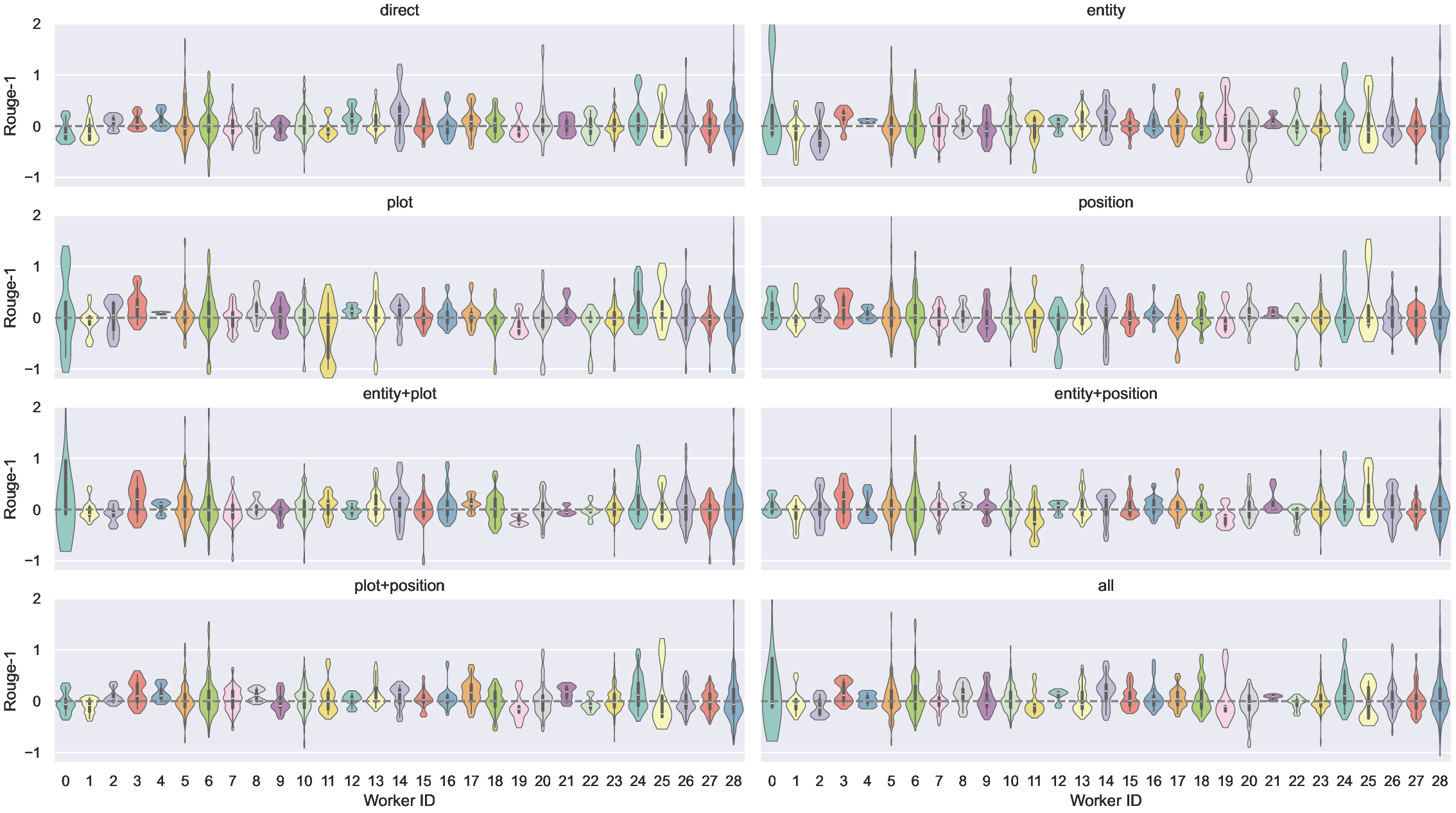, width=400pt, height=210pt}
\vspace{-0.9\baselineskip} 
\caption{Violin plot showing improvements in the ROUGE-1 score from personalized prompting compared to generic summaries using GPT-3.5 Turbo for each worker. The X-axis represents worker IDs, and the Y-axis represents the ROUGE-1 score improvements for PersonalSum dataset.} \label{fig:violin_plot}
\end{figure}

\begin{figure}
\centering
\epsfig{figure=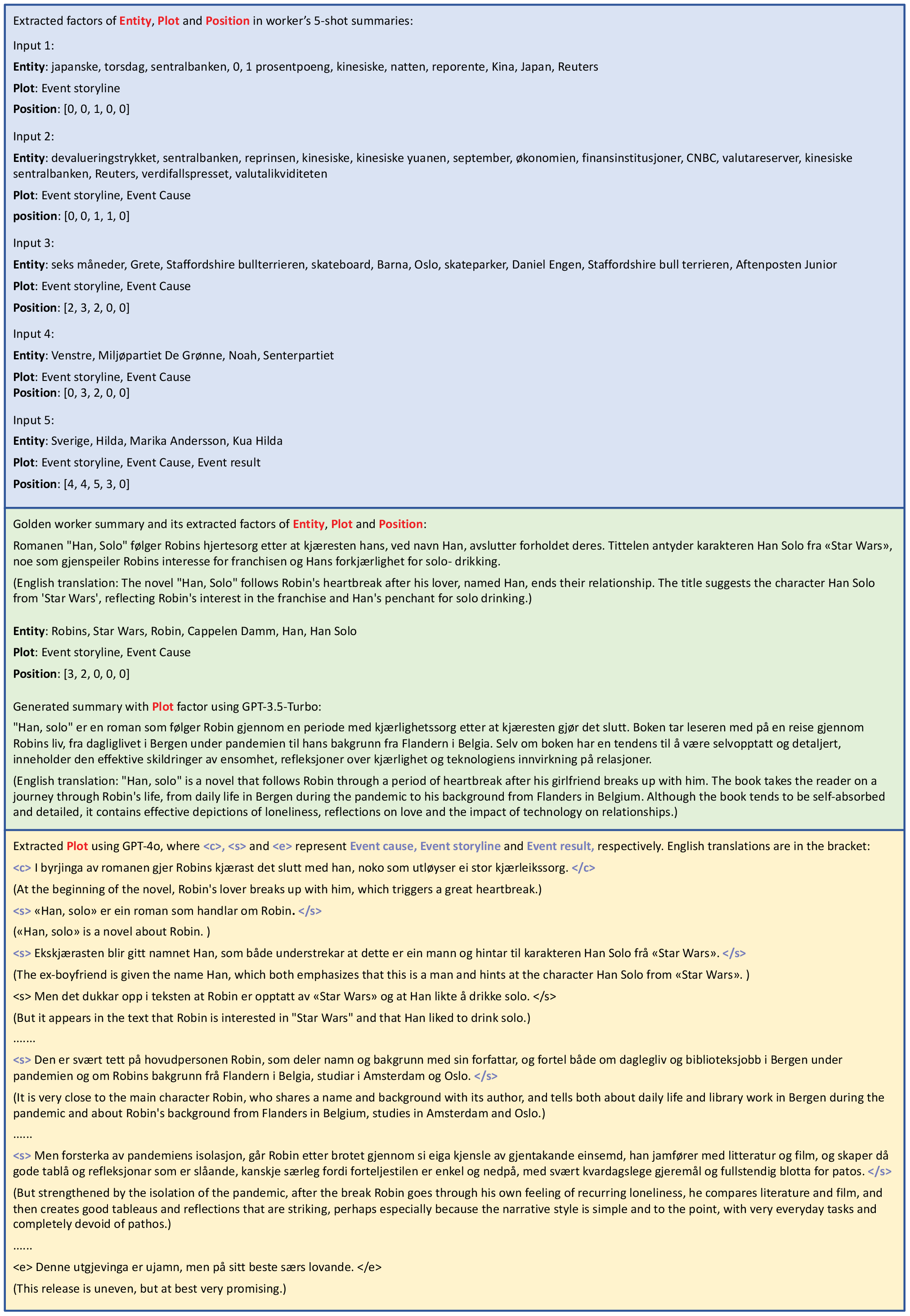, width=400pt}
\caption{The blue box shows the entity, plot, and position factors extracted from the user's five most recent annotation histories. The green box displays the user's current annotation summary, the corresponding extracted factors, and a generated result from GPT-3.5-Turbo with the plot factor. The yellow box shows part of the plot components extracted by GPT-4o related to the generated summary and the worker's annotated summary. (The corresponding English translation is in brackets.)} \label{fig:error_case}
\end{figure}

Figure \ref{fig:violin_plot} shows the improvements for each worker using personalized prompt methods compared to prompting for generating generic summaries. We can observe that some workers, such as workers 3 and 12, show significant improvements with personalized prompting, while others, such as workers 1 and 19, exhibit the opposite trend. To understand the reason for the performance degradation, we selected a case from worker 1 for analysis, as shown in Figure \ref{fig:error_case}. We found that the current article lacks entities related to the worker's previous annotations, and the position information also differs. However, the worker is interested in the storyline and event cause, which has been captured by GPT-3.5-Turbo. The worker's interest is concentrated at the beginning of the article, while the generated summary covers most of it. This issue also appears in other error cases, making it difficult to capture the user's historical interests. One reason is that users cannot choose the articles they want to annotate. This limitation motivated our design to collect Topic-centric PersonalSum.

\begin{figure}
\centering
\epsfig{figure=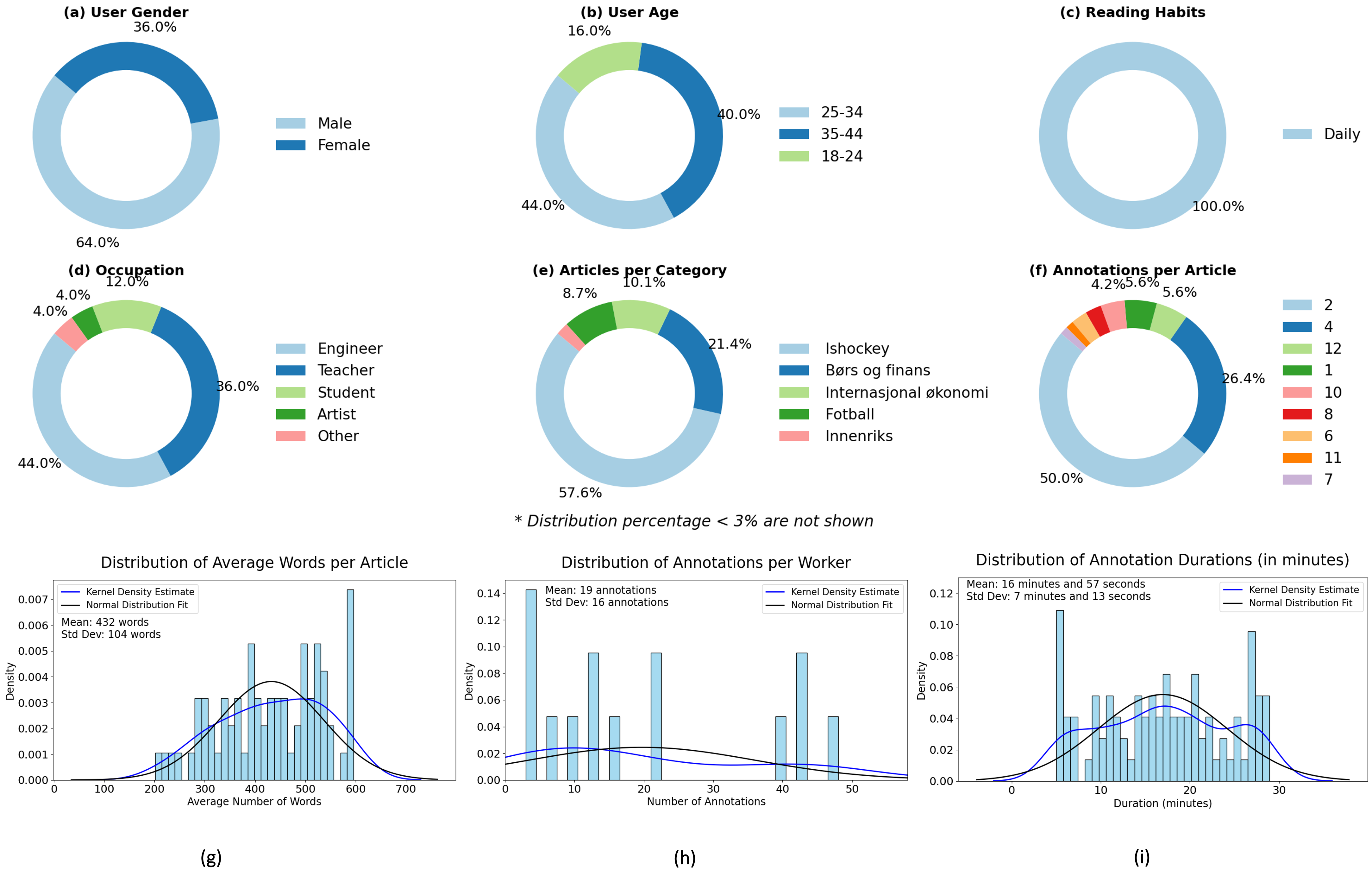, width=385pt, height=229pt}
\caption{Statistics of the annotators, news articles and summaries for Topic-centric PersonalSum.} \label{fig:supplementary_bench}
\end{figure}

\begin{figure}
\centering
\epsfig{figure=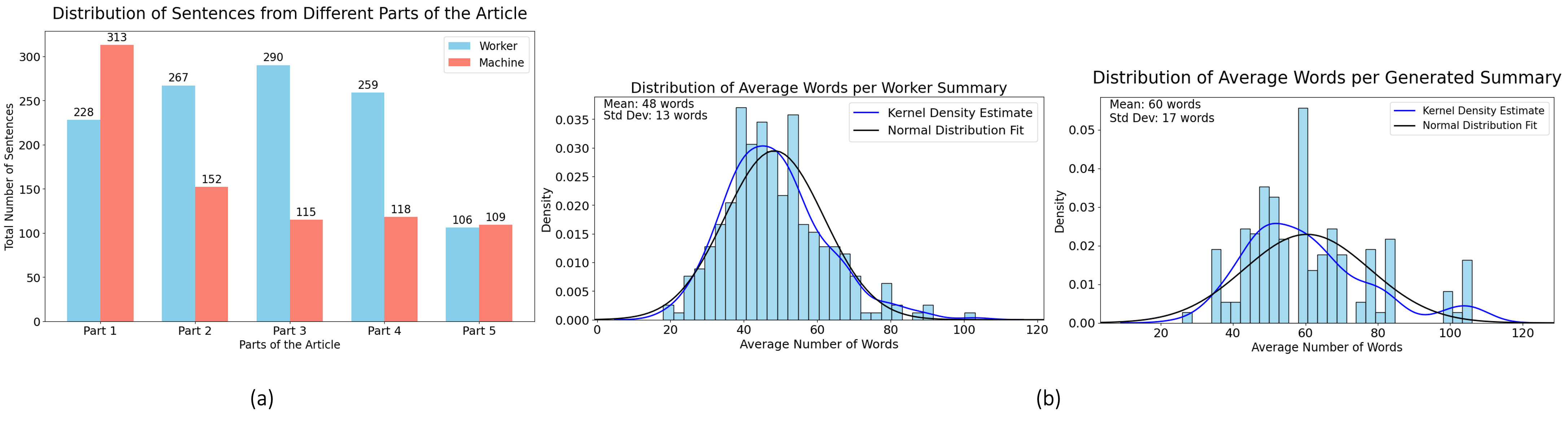, width=395pt, height=90pt}
\caption{(a) The distribution of sources of machine-generated summaries and human-annotated personalized summaries in the Topic-centric PersonalSum. (b) The distribution of average words per machine-generated summary and human-annotated summary in the Topic-centric PersonalSum.} \label{fig:supplementary_difference}
\end{figure}

\subsection{Statistics for Topic-centric PersonalSum}
Topic-centric PersonalSum includes a total of 276 personalized summaries from 72 news articles annotated by 14 distinct Amazon Turkers. Specifically, Figure \ref{fig:supplementary_bench} shows the basic statistics of annotators, news articles and summaries for Topic-centric PersonalSum. Figure \ref{fig:supplementary_difference} shows the distributions of sources and average word counts per summary for machine-generated and human-annotated personalized summaries in Topic-centric PersonalSum.

\begin{figure}
\centering
\epsfig{figure=pictures/personalsum_human_eval.png, width=400pt, height=360pt}
\caption{Instructions for the human evaluation of personalized summaries generated by LLMs.} \label{fig:Instructions}
\end{figure}

\end{document}